\documentclass[10pt,twocolumn,letterpaper]{article}
\usepackage{cuted}
\usepackage{cvpr}              % To produce the CAMERA-READY version
\usepackage{CJKutf8}
\usepackage{float}

\definecolor{cvprblue}{rgb}{0.21,0.49,0.74}
\usepackage[pagebackref,breaklinks,colorlinks,citecolor=cvprblue]{hyperref}
\usepackage{amssymb}
\usepackage{amsmath}
\usepackage{booktabs}
\usepackage{graphicx}
\usepackage{tabularx} 
\usepackage{multirow}
\usepackage{xcolor}

\def\paperID{1898} % *** Enter the Paper ID here
\def\confName{CVPR}
\def\confYear{2025}

\title{DiffSim: Taming Diffusion Models for Evaluating Visual Similarity}

\author{
Yiren Song\thanks{Equal contribution.} \quad Xiaokang Liu\footnotemark[1] \quad Mike Zheng Shou\thanks{Corresponding author.} \\
Show Lab, National University of Singapore \\
}

\begin{document}
\maketitle

\begin{strip}
\centering
    \includegraphics[width=\linewidth]{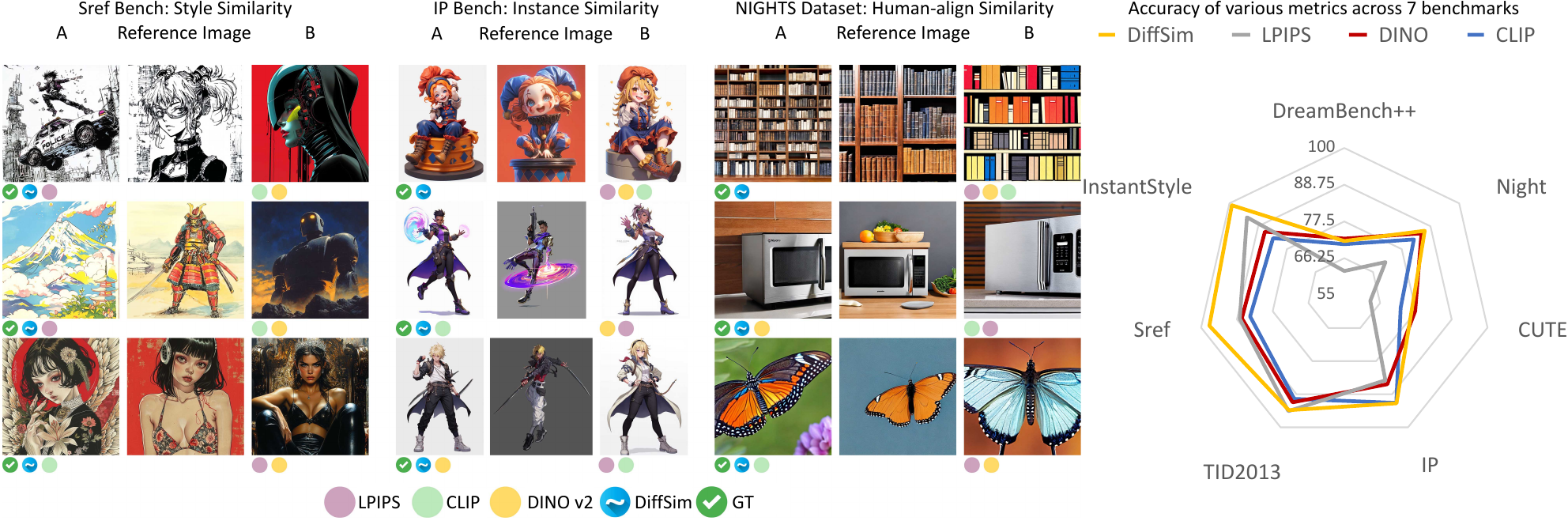}
    \vspace{-7pt}
    \vspace{-0.2cm}
    \captionof{figure}{We propose DiffSim, a method that utilizes pre-trained diffusion models to extract image features for evaluating visual similarity. Our method leads in human judgment consistency, style similarity, and instance-level consistency.}
    \label{fig:teaser}
\end{strip}

\begin{abstract}
Diffusion models have fundamentally transformed the field of generative models, making the assessment of similarity between customized model outputs and reference inputs critically important. However, traditional perceptual similarity metrics operate primarily at the pixel and patch levels, comparing low-level colors and textures but failing to capture mid-level similarities and differences in image layout, object pose, and semantic content. Contrastive learning-based CLIP and self-supervised learning-based DINO are often used to measure semantic similarity, but they highly compress image features, inadequately assessing appearance details. This paper is the first to discover that pretrained diffusion models can be utilized for measuring visual similarity and introduces the DiffSim method, addressing the limitations of traditional metrics in capturing perceptual consistency in custom generation tasks. By aligning features in the attention layers of the denoising U-Net, DiffSim evaluates both appearance and style similarity, showing superior alignment with human visual preferences. Additionally, we introduce the Sref and IP benchmarks to evaluate visual similarity at the level of style and instance, respectively. Comprehensive evaluations across multiple benchmarks demonstrate that DiffSim achieves state-of-the-art performance, providing a robust tool for measuring visual coherence in generative models. Code is released at \href{https://github.com/showlab/DiffSim}{https://github.com/showlab/DiffSim}.
\end{abstract}    
\vspace{-1.5em}
\section{Introduction}

Assessing the consistency of appearance and style between customized generation results and reference images is an important issue.  Some studies \cite{dreambench++, ssr} resort to time-consuming user studies as a supplementary evaluation, despite questions about their fairness and objectivity. Existing evaluation metrics such as CLIP-Image score  \cite{clip} and DINO score \cite{dinov1}, while commonly used, often fail to reflect subtle differences due to their reliance on high-dimensional semantic features for cosine similarity computation. Research \cite{dreamsim} indicates that these metrics sometimes do not align with human subjective evaluations and are insufficient for comprehensively measuring consistency in appearance and style details in generation tasks. Some studies propose human-aligned similarity assessment methods \cite{dreamsim}, which collect data on human choices of similarity triplets and train models, but their generalization ability in out-of-domain scenarios is considered limited. To overcome these challenges, there is an urgent need for more effective image similarity evaluation metrics in the image generation field.
        
The motivation of this paper is to assess visual similarity using pre-trained diffusion models. Our rationale is based on three key insights: 1. ReferenceNet \cite{referenceonly, animateanyone, magicanimate}, uses U-Net to extract features from reference images and directly concatenates these features (K map, V map) in the denoising U-Net's self-attention layer, effectively maintaining appearance similarity. 2. Custom diffusion \cite{customdiffusion} demonstrates that the to K and to V matrices in the cross-attention layer are critical modules for stable diffusion models to learn concepts. 3. IP-Adapter \cite{ipa} injects IP tokens into the cross-attention layer, achieving consistent imagery generation. Since the features of these layers are crucial for generating samples with consistent visual appearance, we believe these features represent the visual appearance and concepts and can be used to assess visual similarity. Recent studies have utilized pre-trained diffusion models for perception tasks \cite{dift, general, probing}, demonstrating the potential of generative models in perception tasks and the universality of diffusion model features.

However, two design challenges arise: (1) U-Net, trained on noisy images for denoising purposes, requires noise to be added to the input images (simulating forward diffusion) before feature extraction. (2) Unlike CLIP and DINO, which use high-dimensional semantic features for cosine similarity, Stable Diffusion U-Net's features are densely packed with spatial information, leading to misalignment at the pixel level. Therefore, simple MSE or cosine calculations between feature maps are impractical, as validated by our subsequent experiments. To address this, we introduce the Aligned Attention Score (AAS), which innovatively uses attention mechanisms to align features of images \(A\) and \(B\) in the self-attention layer of U-Net, and then calculates the cosine distance between the aligned features. 

Furthermore, we explore the variations in features across different layers and denoising timesteps within the denoising U-Net. We find that shallower layers and higher denoising timesteps are suitable for evaluating low-level and style similarities, while deeper layers and lower timesteps excel at assessing semantic similarity. This implies that DiffSim can achieve different similarity measurements with simple configuration adjustments. Besides calculating similarity in the self-attention layers, we also explored using IPAdapter-Plus in the cross-attention layers to evaluate visual similarity. Additionally, we discovered that this technique can be generalized to enhance other architectures such as CLIP and DINO, introducing the CLIP AAS metric and DINO ASS metric, which significantly improved performance in certain tasks.

Extensive evaluations across multiple benchmarks have proven the effectiveness and advancement of DiffSim. Despite its seemingly straightforward approach, DiffSim excels in all tasks without any additional fine-tuning or supervision, surpassing both CLIP and DINO v2. Remarkably, our experiments reveal that DiffSim's assessments of image similarity are highly consistent with human judgments, ranking at the forefront in two human consistency benchmarks. Furthermore, DiffSim performs exceptionally well on synthetic images, achieving the best results in style similarity and appearance consistency assessments on our newly proposed Style-ref and IP-ref benchmarks.

We summarize our main contributions as follows:
\begin{itemize}
    \item We introduce DiffSim, an innovative image similarity assessment method that utilizes the denoising U-Net of pre-trained diffusion models to evaluate visual similarity without the need for additional fine-tuning or specific data supervision.
    \item We propose the Aligned Attention Score, which precisely aligns image features through attention mechanisms, effectively addressing alignment and information loss issues in traditional assessments, and enables multi-dimensional similarity measurements based on the characteristics of different layers and denoising timesteps in diffusion models.
    \item We introduce two new benchmarks—Sref and IP bench—to assess style and instance consistency. Extensive experiments and evaluations demonstrate the effectiveness and advancement of DiffSim, which not only surpasses existing CLIP and DINO models but also aligns closely with human visual assessment.
\end{itemize}

\vspace{-0.5em}
\section{Related Work}

\subsection{Diffusion Models}
Diffusion probability models \citep{ddim,ddpm} are advanced generative models that restore original data from pure Gaussian noise by learning the distribution of noisy data at various levels of noise. Their powerful capability to adapt to complex data distributions has led diffusion models to achieve remarkable success across several domains including image synthesis \citep{rombach2022high,dit}, image editing \citep{instructpix2pix,p2p,stablemakeup,stablehair}, and video gneration \citep{animatediff, svd, processpainter}. Stable Diffusion \citep{rombach2022high} (SD), a notable example, utilizes a U-Net architecture and extensively trains on large-scale text-image datasets to iteratively generate images with impressive text-to-image capabilities. Enhancements in controllable image generation have been driven by methods such as ControlNet \citep{controlnet} and T2I-adapter \citep{t2i}, which significantly improve controllability over generated images by employing multimodal inputs such as depth maps and segmentation maps.

Customized generation methods  enable flexible customization of concepts and styles by fine-tuning U-Net \citep{dreambooth} or certain parameters \citep{lora, customdiffusion}, alongside trainable tokens. Training-free customization methods \citep{ipa, instantid, instantstyle, ssr, fast} leverage pre-trained CLIP \citep{clip} or Arcface \citep{arcface} encoders to extract image features. These features are then injected into the cross-attention layers of U-Net through adapter structures for efficient generation. Methods like ReferenceNet \citep{animateanyone, magicanimate} integrate reference image features into the self-attention layer of a denoising U-Net, showing advantages in maintaining appearance similarity. These are widely applied in tasks such as image editing \citep{stablehair, stablemakeup}, facial and body animation \citep{animateanyone, magicanimate}, and Image2Video tasks \citep{i2vadapter,i2vgen-xl}.

Additionally, recent research has shown that pre-trained diffusion models can also address perception tasks such as 3D awareness \cite{probing}, keypoint matching \cite{dift}, and image classification \cite{chen2023robust}. This paper demonstrates the practicality and technical maturity of using pre-trained diffusion models for image similarity assessment.

\vspace{-0.3em}
\subsection{Perceptual Similarity}
Traditional metrics such as Manhattan \(l_1\), Euclidean \(l_2\), Mean Squared Error (MSE), and Peak Signal-to-Noise Ratio (PSNR) use point-to-point differences to measure similarity, but lack the ability to effectively understand the global structure and content of images. In contrast, deep learning-based metrics that utilize features extracted from pretrained networks like VGG \cite{vgg} and AlexNet \cite{alexnet} have proven superior. Further improvements in these metrics have been achieved through optimizations on perceptual data, as seen in methods like LPIPS \cite{lpips}, PIE-APP \cite{pieapp}, and DreamSim \cite{dreamsim}. Research by Muttenthaler \citep{humanalign} on concept similarity within a subset of the THINGS dataset \cite{things} provides insights into high-level human similarity. 

Meanwhile, DreamSim \cite{dreamsim} integrates features from models like CLIP \cite{clip} and DINO \cite{dinov2}, trained on synthetic datasets and human annotations, to achieve state-of-the-art results. However, CLIP and DINO encode images into 1×768-dimensional feature vectors for cosine similarity computation, emphasizing categorical attributes over fine-grained appearance similarity. This high compression sometimes leads to divergences from human perception, indicating a need for more effective visual similarity metrics in the generative AI and metric learning fields.

% 传统的度量方法（曼哈顿$l_1$、欧几里得$l_2$、均方误差(MSE)和峰值信噪比(PSNR)）采用点对点的差异来衡量相似性， 缺乏全局感知能力， 而无法有效理解图像的整体结构和内容。基于深度学习的度量利用从VGG或AlexNet等预训练网络提取的特征，表现优于传统度量。进一步在感知数据上的优化，如LPIPS、PIE-APP和DPAM等，已证明可以提升度量效果。此外，为提高鲁棒性采用集成方法，为增强稳定性使用抗锯齿技术，以及采用全局描述符改善纹理表现，都有助于改进性能。Muttenthaler等人的研究通过在THINGS数据集的子集上专注于概念相似性，提供了对高级人类相似性的见解。

% 利用在大规模数据集上以自监督方式训练的深度模型学到的特征（CLIP，DINO），被广泛用于衡量图像相似度和客制化生成的一致性。 然而CLIP，DINO将图像压缩到1*768的特征并计算cos similarity, 信息高度压缩，更关注类别属性而非外观相似度， 之前的研究表明CLIP和DINO的判断有时和人类感知结果不同。DreamSim通过收集合成数据集和人类标注，集成CLIP和DINO等模型的特征进行训练，取得sota效果。综上，生成式AI社区和Metric learning领域急需一种更好的图像一致性指标。

\vspace{-0.3em}
\subsection{Style Similarity Evaluation}

% 艺术风格通常被定义为与艺术家或艺术运动相联系的图像的全局特征集合，这包括颜色使用、笔触技巧、构图及透视等元素。早期算法通过低级视觉特征来建模风格，而现代方法更多采用神经网络处理风格转移与分类任务。如Gatys等人采用Gram矩阵作风格描述符，Luan等人则添加光照真实性正则化项优化风格转移。最近，Wang等人开发了基于合成风格对的风格识别模型，而CSD项目则通过实际图像对训练模型。

% 在生成式AI的背景下，LoRA模型等技术通过训练少量风格图像生成风格一致的新样本，或者通过IPA、Instant-Style等方法直接获得风格表征。结合文本提示，上述技术可以创造全新的艺术风格，从而扩展了风格的定义并使其更加多样化和模糊。这种风格的多样性和新颖性给风格相似性评估带来了挑战，要求评估方法能更精细地捕捉和理解图像间的风格关系。为此，我们提出的DiffSim方法利用大规模图文对上预训练的扩散模型，无需任何微调便实现了先进的风格相似性评估。

Artistic style is often defined as a collection of global characteristics associated with an artist or artistic movement, encompassing elements like color usage, brushstroke techniques, composition, and perspective. Early algorithms modeled style using low-level visual features \cite{sablatnig1998, hughes2011}, while modern approaches predominantly utilize neural networks for style transfer and classification tasks. For instance, Gatys et al. \cite{gatys2016} utilized Gram matrices as style descriptors, while Luan et al. \cite{luan2017deepphotostyletransfer} introduced a photorealism regularization term to optimize style transfer. Recently, Wang et al. \cite{wang2023evaluating} developed a style recognition model based on synthetic style pairs, and CSD \cite{somepalli2024measuring} introduces a multi-label contrastive learning scheme to extract style descriptors.

In the era of generative AI, techniques like the LoRA model \cite{hu2021loralowrankadaptationlarge} can generate new samples consistent with a few trained style images, or methods like IP-Adapter \cite{ye2023ip} and InstantStyle \cite{instantstyle} directly obtain style representations. When combined with text prompts, these technologies can create entirely new artistic styles, adding complexity and ambiguity. This diversity in style poses challenges for style similarity evaluation, demanding methods that more finely capture and understand stylistic relationships between images. To address this, our proposed DiffSim method leverages the diffusion model trained on massive image-text pairs, enabling advanced style similarity assessment without any fine-tuning.

\begin{figure*}[htp]
    \centering
    \includegraphics[width=1.0\linewidth]{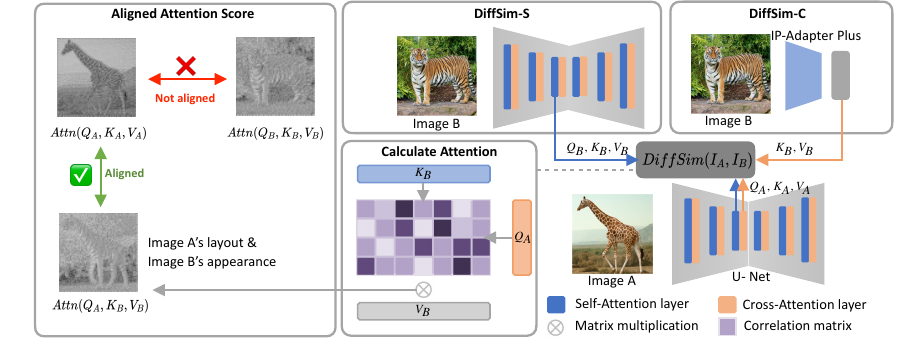}
    \caption{The illustration shows two DiffSim implementations: DiffSim-S using self-attention, where U-Net extracts features from both images to compute Aligned Attention Score (AAS) at a specified layer; and DiffSim-C using cross-attention, where features are extracted via IP-Adapter Plus and U-Net with swapped image inputs.}
    \label{fig:method}
\end{figure*}

% 这张示意图展示了用自注意力层和交叉注意力层特征计算DiffSim的两种实现。当在自注意力计算时，我们通过U-Net分别提取两张图片的特征，并在特定层计算Aligned Attention Score（AAS）。 当在交叉注意力层计算时时，我们用IP-Adapter Plus和U-Net提取特征，并交换两张图像的输入位置。

% \begin{figure*}
%   \centering
%   \begin{subfigure}{0.68\linewidth}
%     \fbox{\rule{0pt}{2in} \rule{.9\linewidth}{0pt}}
%     \caption{An example of a subfigure.}
%     \label{fig:short-a}
%   \end{subfigure}
%   \hfill
%   \begin{subfigure}{0.28\linewidth}
%     \fbox{\rule{0pt}{2in} \rule{.9\linewidth}{0pt}}
%     \caption{Another example of a subfigure.}
%     \label{fig:short-b}
%   \end{subfigure}
%   \caption{Example of a short caption, which should be centered.}
%   \label{fig:short}
% \end{figure*}

% 传统的图像相似性度量，如曼哈顿 \(l_1\)、欧几里得 \(l_2\)、均方误差 (MSE) 和峰值信噪比 (PSNR)，通常关注像素级的差异。这些度量虽然有用，但常常无法包含人类视觉的细微感知，特别是在理解复杂的图像内容或结构模糊性时。另一方面，基于补丁的度量如结构相似性指数 (SSIM)、特征相似性指数 (FSIM) 和高动态范围视觉差异预测器 (HDR-VDP-2) 在解决光度失真方面取得了显著进展，但在面对更复杂的视觉任务时仍显不足。随着深度学习的兴起，度量标准已向利用深度神经网络的特征空间转变。这些基于学习的度量利用从大量图像数据集上训练的模型中提取的深度特征，提供了更符合人类感知判断的相似性评估。这种范式转变导致了如学习感知图像补丁相似性 (LPIPS) 和深度感知相似性度量 (DISTS) 等先进的感知度量的发展，这些方法在度量图像相似性时并不纯粹，过多的考虑layout等信息，即时同一张图像经过几个像素的错位，也会计算出较高的LPIPS。 此外，一些方法利用自监督学习或对比学习的模型实现相似性的度量，例如CLIP， DINO，DINO v2。

\vspace{-0.5em}
\section{Method}
%本节我介绍DiffSim的方法。 3.1节介绍包括扩散模型和Stable diffusion model 中的注意力机制作为前置知识。 3.2节介绍DiffSim核心实现方法 3.3节介绍如何在不同模型架构上实现DiffSim，扩散稳定扩散模型、CLIP、DINO。

 % Section Sec. \ref{sec:1} presents attention mechanisms in the Stable Diffusion model. Sec. \ref{sec:2} introduce Aligned Attention Score we proposed. Sec. \ref{sec:3} describes how to implement DiffSim across different model architectures. Sec. \ref{sec:4} introduce two benchmark we propose.

DiffSim leverages the implicit alignment mechanism of attention to align two images at the style or semantic level, followed by similarity computation. In Sec. \ref{sec:1}, we briefly review the attention mechanism in the Stable Diffusion model. Sec. \ref{sec:2} introduces the mechanism for calculating AAS (Aligned Attention Score) within a single attention layer. Sec. \ref{sec:3} provides a detailed explanation of how DiffSim is computed using pretrained models and AAS. Sec. \ref{sec:4} introduce two benchmark we propose.

\vspace{-0.3em}
\subsection{Preliminaries}
\label{sec:1}

\subsubsection{Cross Attention in Diffusion Model} 
Taking Stable Diffusion as an example, cross-attention layers play a crucial role in fusing images and texts, allowing diffusion models to generate images that are consistent with textual descriptions. The cross-attention layer receives the query, key, and value matrices, i.e., $Q_{\text{cross}}$, $K_{\text{cross}}$, and $V_{\text{cross}}$, from the noisy image and prompt. Specifically, $Q_{\text{cross}}$ is derived from the spatial features of the noisy image $\phi_{\text{cross}}(z_t)$ by learned linear projections $\ell_q$, while $K_{\text{cross}}$ and $V_{\text{cross}}$ are projected from the textual embedding $P_{\text{emb}}$ of the input prompt $P$ using learned linear projections denoted as $\ell_k$ and $\ell_v$, respectively. The cross-attention map is defined as:
\begin{align}
Q_{\text{cross}} &= \ell_q(\phi_{\text{cross}}(z_t)), \\
K_{\text{cross}} &= \ell_k(P_{\text{emb}}), \\
M_{\text{cross}} &= \text{Softmax}\left(\frac{Q_{\text{cross}} K_{\text{cross}}^T}{\sqrt{d_{\text{cross}}}}\right),
\end{align}
where $d_{\text{cross}}$ is the dimension of the keys and queries. The final output is defined as the fused feature of the text and image, denoted as $\hat{\phi}(z_t) = M_{\text{cross}} V_{\text{cross}}$, where $V_{\text{cross}} = \ell_v(P_{\text{emb}})$. Intuitively, each cell in the cross-attention map, denoted as $M_{ij}$, determines the weights attributed to the value of the $j$-th token relative to the spatial feature $i$ of the image. The cross-attention map enables the diffusion model to locate/align the tokens of the prompt in the image area.

% 如图2所示，与交叉注意力不同，自注意力层通过学习的线性投影 $\bar{\ell}K$ 和 $\bar{\ell}Q$ 分别从噪声图像 $\phi{\text{self}} (z_t)$ 接收键矩阵 $K{\text{self}}$ 和查询矩阵 $Q_{\text{self}}$。自注意力图定义如下： \begin{align} Q_{\text{self}} &= \bar{\ell}q(\phi{\text{self}} (z_t)), \ K_{\text{self}} &= \bar{\ell}K(\phi{\text{self}} (z_t)), \ M_{\text{self}} &= \text{Softmax}\left(\frac{Q_{\text{self}} K_{\text{self}}^T}{\sqrt{d_{\text{self}}}}\right), \end{align} 其中 $d_{\text{self}}$ 是 $K_{\text{self}}$ 和 $Q_{\text{self}}$ 的维度。$M_{\text{self}}$ 决定了分配给图像中第 $i$ 和第 $j$ 个空间特征相关性的权重，并且可以影响生成图像的空间布局和形状细节。因此，自注意力图可以用来在图像编辑过程中保持原始图像的空间结构特征

\subsubsection{Self Attention in Diffusion Model}  Unlike cross-attention, the self-attention layer receives the keys matrix $K_{\text{self}}$ and the query matrix $Q_{\text{self}}$ from the noisy image $\phi_{\text{self}} (z_t)$ through learned linear projections $\bar{\ell}_K$ and $\bar{\ell}_Q$, respectively. The self-attention map is defined as:
\begin{align}
Q_{\text{self}} &= \bar{\ell}_q(\phi_{\text{self}} (z_t)), \\
K_{\text{self}} &= \bar{\ell}_K(\phi_{\text{self}} (z_t)), \\
M_{\text{self}} &= \text{Softmax}\left(\frac{Q_{\text{self}} K_{\text{self}}^T}{\sqrt{d_{\text{self}}}}\right),
\end{align}
where $d_{\text{self}}$ is the dimension of $K_{\text{self}}$ and $Q_{\text{self}}$. $M_{\text{self}}$ determines the weights assigned to the relevance of the $i$-th and $j$-th spatial features in the image and can affect the spatial layout and shape details of the generated image. Consequently, the self-attention map can be utilized to preserve the spatial structure characteristics of the original image throughout the image editing process.

\vspace{-0.2em}
\subsection{Aligned Attention Score}
\vspace{-0.2em}
\label{sec:2}

In this section, we introduce the Aligned Attention Score (AAS), a new metric designed to provide implicit alignment between images while measuring various aspects such as semantic content and style. Traditional similarity assessment methods, such as Mean Squared Error (MSE) or cosine similarity, often assume that the latent features of images are pixel-aligned. However, this assumption frequently fails in practical applications due to significant changes in style, pose, or context, leading to poor and inaccurate similarity measurements. Moreover, popular methods like CLIP and DINO compress high-dimensional block features into a lower-dimensional feature space using Multilayer Perceptrons (MLPs) and compute cosine similarity on these simplified representations. While effective at semantic-level comparisons, this compression process can result in the loss of critical detail, failing to capture subtle differences vital for some applications.

AAS addresses misalignment issues by leveraging attention mechanisms in pretrained U-Net or Transformer-based models. We define \(L_A\) and \(L_B\) as the latent representations of images \(I_A\) and \(I_B\) respectively. AAS dynamically aligns these representations using the neural network’s attention function \( \text{attn}(Q, K, V) \):
\begin{align}
\text{AAS}(L_A, L_B) &= \cos(\text{attn}(Q_A, K_A, V_A), \text{attn}(Q_A, K_B, V_B)), \label{eq:aas1} \\
\text{AAS}(L_B, L_A) &= \cos(\text{attn}(Q_B, K_B, V_B), \text{attn}(Q_B, K_A, V_A)). \label{eq:aas2}
\end{align}
By aligning dense latent representations in the attention layers, AAS ensures that each image’s features are evaluated both as queries and keys against the features of the other image. This method not only compensates for the lack of pixel alignment but also preserves the richness of the feature space, providing a more accurate method of measuring perceptual similarity that closely aligns with human visual judgment:
\begin{equation}
\text{Similarity}(L_A, L_B) = \text{AAS}(L_A, L_B) + \text{AAS}(L_B, L_A). \label{eq:similarity}
\end{equation}
This integrated approach to image alignment and similarity assessment leverages different granularity representations provided by pretrained models, effectively improving the handling of image misalignment and information loss.

%subsection{对齐注意力分数}
% 在本节中，我们介绍了对齐注意力分数（AAS），这是一种旨在克服传统相似性评估（如均方误差（MSE）或余弦相似度）的局限性的新指标。这些传统方法假设图像潜在特征之间在像素层面上是对齐的，但在实际应用中，图像常常因风格、姿态或上下文的显著变化而违反这一假设，导致相似性测量的效果不佳和不准确。

% 此外，如CLIP和DINO等流行方法通过多层感知机（MLPs）将高维块特征压缩到低维特征空间，并在这些简化的表征上计算余弦相似度。尽管这种方法在语义级比较上有效，但压缩过程会导致大量细节信息的丢失，未能捕捉到某些应用中至关重要的微妙差别。

% 我们定义 \(L_A\) 和 \(L_B\) 分别为图像 \(I_A\) 和 \(I_B\) 的潜在表征。为了解决错位和信息丢失的问题，我们的实现通过神经网络中的注意力函数 $\text{attn}(Q, K, V)$ 动态对齐这些表征：\begin{align}   \text{AAS}(L_A, L_B) &= \cos(\text{attn}(Q_A, K_A, V_A), \text{attn}(Q_A, K_B, V_B)) \\\text{AAS}(L_B, L_A) &= \cos(\text{attn}(Q_B, K_B, V_B), \text{attn}(Q_B, K_A, V_A))\end{align}通过在预训练的U-Net或基于Transformer的模型中的注意力机制处理特征，AAS在注意力层中对齐两个密集的潜在表示，确保每个图像的特征既作为查询也作为键与另一图像的特征进行评估，从而提供了一种健壮的视觉相似性测量方法：\begin{equation}\text{Similarity}(L_A, L_B) = \text{AAS}(L_A, L_B) + \text{AAS}(L_B, L_A)\end{equation}这种方法不仅补偿了缺乏像素对齐的问题，还保留了特征空间的丰富性，提供了更准确地反映与人类视觉判断密切相符的感知相似性的方法。

%在本节中，我们介绍了对齐注意力分数（AAS），这是一种新的相似性评估指标，设计用来隐式地对齐图像，并在语义和风格等不同层面上进行测量。传统的相似性评估方法，如均方误差（MSE）或余弦相似度，通常假设图像的潜在特征在像素层面上是对齐的。然而，在实际应用中，由于风格、姿态或上下文的显著变化，这一假设往往不成立，导致相似性测量的效果不佳和不准确。此外，流行的方法如CLIP和DINO通过多层感知机（MLPs）将高维块特征压缩到低维特征空间，并在这些简化的表征上计算余弦相似度。尽管这种方法在语义级比较上有效，但压缩过程可能导致大量细节信息的丢失，未能捕捉到某些应用中至关重要的微妙差别。

%AAS 通过在预训练的U-Net或基于Transformer的模型中的注意力机制处理特征来解决错位问题。我们定义 \(L_A\) 和 \(L_B\) 分别为图像 \(I_A\) 和 \(I_B\) 的潜在表征。AAS 使用神经网络中的注意力函数 $\text{attn}(Q, K, V)$ 动态对齐这些表征：

%通过在注意力层中对齐两个密集的潜在表示，AAS 确保每个图像的特征既作为查询也作为键与另一图像的特征进行评估。这种方法不仅补偿了缺乏像素对齐的问题，还保留了特征空间的丰富性，提供了一种更准确地反映与人类视觉判断密切相符的感知相似性的方法：

%这种对图像进行对齐和相似性评估的集成方法利用了预训练模型提供的不同粒度的表征，有效提升了对图像错位和信息丢失问题的处理能力。

\vspace{-0.3em}
\subsection{DiffSim Metric}
\vspace{-0.2em}
\label{sec:3}
To fully utilize the attention mechanisms in stable diffusion, we repurpose its self-attention and cross-attention for our AAS, forming two methods: DiffSim-S and DiffSim-C.

\subsubsection{DiffSim-S}
\label{sec:diffsim_self}
% 在海量图像文本对上预训练的stable diffusion U-Net是出色的视觉特征提取器，reference only方法直接将参考图像输入U-Net并将自注意力层中参考图的K和V直接与去噪U-Net的自注意力层中的K和V相连，实现了一致性的图像variation。 类似的，人体驱动任务中animate anyone, magicanimate等将U-Net作为提取参考图像细节信息的ReferenceNet, 并将ReferenceNet和去噪U-Net的自注意力层相连进行条件信息的注入。同样的做法还被应用于人脸驱动和和Image2Video 。

% 在上述方法中，自注意力层中特征的嵌入可以实现外观一致性的图像或视频的生成，因此我们认为U-Net自注意力层的特征可以反应图像外观的一致性。我们提出了DiffSim的自注意力层实现。具体来说，我们计算

%The pre-trained Stable Diffusion U-Net, extensively trained on massive image-text pairs, excels as a visual feature extractor. The reference only \citep{referenceonly} approach inputs reference images directly into the U-Net, connecting the reference image's keys (K) and values (V) in the self-attention layers directly with those in the denoising U-Net's self-attention layers to achieve consistent image variations. Similarly, in human-driven tasks such as Animate Anyone and MagicAnimate \citep{magicanimate, animateanyone}, the U-Net serves as a ReferenceNet to extract detailed information from reference images, linking its self-attention layers with those of the denoising U-Net for injecting conditional information. This technique is also applied in face-driven tasks \citep{emo, echomimic, idanimator, x-portrait}, image customization generation \citep{referenceonly, stablemakeup, stablehair}, and Image2Video generation \citep{livephoto, i2vgen-xl}. %

It is well known that the features in the U-Net’s self-attention layers can reflect the consistency of image appearance. Previous works \cite{magicanimate, animateanyone, emo, echomimic, idanimator, x-portrait, referenceonly, stablemakeup, stablehair, livephoto, i2vgen-xl} have utilized this characteristic to achieve high-consistency image and video generation.

In the aforementioned methods, the embedding of features in the self-attention layers enables the generation of images or videos with appearance consistency. Thus, we posit that the features in the U-Net's self-attention layers can reflect the consistency of image appearance. We propose an implementation of the self-attention layers in DiffSim. We caculate:
\begin{equation}
\begin{aligned}
\text{DiffSim-S}(I_A, I_B, n, t) &= \frac{1}{2} \left( \text{AAS}(z_{t, \text{self}, n}^A , z_{t, \text{self}, n}^B ) \right.  \\
&\quad \left. + \text{AAS}(z_{t, \text{self}, n}^B , z_{t, \text{self}, n}^A ) \right),
\end{aligned}
\end{equation}
where \(z_{t, \text{self}, n}^A\) and \(z_{t, \text{self}, n}^B\) denote the latent representations of images \(I_A\) and \(I_B\) within the \(n\)-th self-attention layer of the U-Net at denoising timestep \(t\), respectively.

\begin{figure*}[htp]
    \centering
    \includegraphics[width=1.0\linewidth]{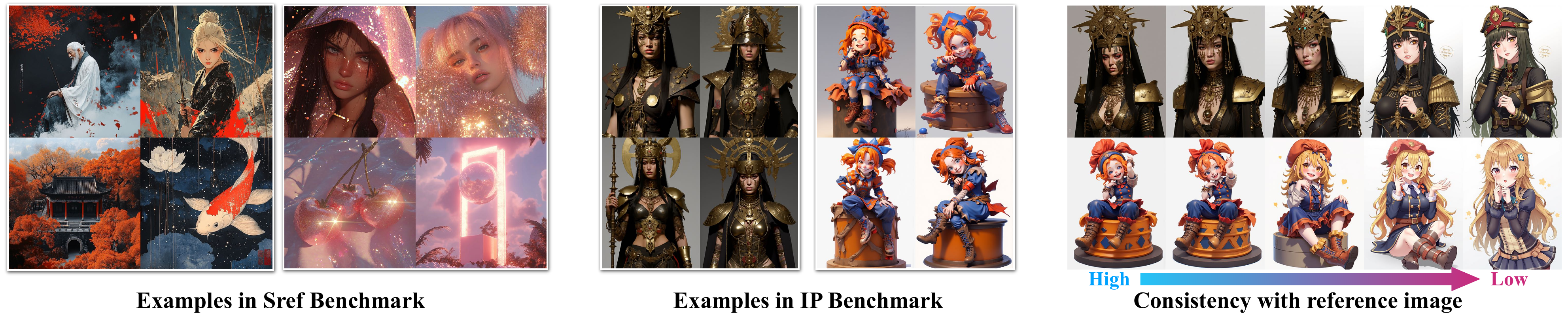}
    \caption{To evaluate style similarity and instance-level similarity, we introduced the Sref bench and IP bench. The Sref dataset contains 508 styles, each generated by Midjourney's sref mode and handpicked by human artists, represented through four different thematic reference images. The IP dataset includes a set of 299 IPs comprising highly similar images along with variants that gradually decrease in similarity.}
    \label{bench}
\end{figure*}

% 为了评估风格相似性和instance-level similarity， 我们提出了Sref dataset 和 IP dataset。Sref dataset包含508种风格，每种风格都是由Midjourney的sref模式生成然后人类艺术家手工挑选，并通过四张不同主题的参考图片来展示。IP dataset 包含299个IP 的一组高度相似的图像，和其相似度逐渐降低的变体。

\vspace{-0.6em}
\subsubsection{DiffSim-C}
\vspace{-0.2em}
The Stable diffusion enhances text-to-image generation by integrating text embeddings as conditions within its cross-attention layers. Prior studies \cite{customdiffusion, appearance} have highlighted the pivotal role of these layers in learning customized model concepts, demonstrating that new concepts can be learned through mere fine-tuning of these layers. Given this context, we explored the potential of cross-attention layer features to assess image similarity. However, traditionally, cross-attention layers only support the calculation of attention between image latents and text embeddings. To extend their capability, we employed IP-Adapter Plus \cite{ipa}, which utilizes patch features from the CLIP image encoder and increases the number of IP tokens to 16, injecting these tokens into the cross-attention layers of U- Net. This setup facilitates a more nuanced computation of attention that involves interaction between image latents and enhanced IP tokens.

We designed the structure illustrated in Figure ~\ref{fig:method}, where Image \(I_A\) and Image \(I_B\) are input into IP-Adapter Plus and the denoising U-Net, respectively. The attention-based similarity scores are then computed by comparing the latents of one image to the IP tokens derived from the other, and vice versa. Specifically, we calculate:
\begin{equation}
\begin{aligned}
\text{DiffSim-C}(I_A, I_B, n, t) &= \frac{1}{2} \left( \text{AAS}(z_{t, \text{cross}, n}^A , IP_B) \right. \\
&\quad \quad \left. + z_{t, \text{cross}, n}^B, IP_A) \right),
\end{aligned}
\end{equation}
where \(z_{t, \text{cross}, n}^A\) and \(z_{t, \text{cross}, n}^B\) denote the latent representations of Images \(I_A\) and \(I_B\) within the \(n\)-th cross-attention layer of the U-Net at denoising timestep \(t\), respectively. Similarly, \(IP_A\) and \(IP_B\) represent the IP tokens associated with Images \(I_A\) and \(I_B\), respectively.

\vspace{-0.3em}
\subsubsection{DiffSim Adaptation to Various Frameworks}
% DiffSim的方法具有广泛的适用性，可以通过计算注意力层中特征相似性来分析两张图像的特征。我们已将DiffSim方法扩展应用于CLIP Image Encoder和DINO v2架构。CLIP通过对比学习训练图像和文本编码器，以理解图像与文本之间的关系，并最大化匹配图文对的相似性，通常采用Vision Transformer或ResNet作为图像编码器。DINO采用自监督方式，通过教师-学生网络架构，其中两者均使用Vision Transformer，学生网络通过模仿经数据增强的输入和教师网络的输出来学习视觉表示，教师网络则通过指数移动平均（EMA）进行更新。在特定自注意力层上，我们计算图像相似度，开发了新的指标CLIP Cross-SiM和DINO v2 Cross-Sim来评估相似性。

The DiffSim method is widely applicable and allows for the analysis of image features by computing feature similarity in the attention layers. We have extended the DiffSim approach to the CLIP Image Encoder and DINO v2 architectures. CLIP trains image and text encoders through contrastive learning to understand the relationship between images and text, aiming to maximize the similarity of matching image-text pairs, typically using Vision Transformer or ResNet as the image encoder. DINO operates in a self-supervised manner with a teacher-student network architecture, both utilizing Vision Transformers. The student network learns visual representations by mimicking the output of the teacher network, which has been augmented with random data transformations. The teacher network is updated through exponential moving average (EMA). For specific self-attention layers in the CLIP and DINO v2 models, we calculate AAS and introduce new metrics, CLIP AAS and DINO v2 AAS.

\vspace{-0.3em}
\subsection{New Benchmarks}
\vspace{-0.2em}
\label{sec:4}
\subsubsection{Sref Bench} Style is subjective, thus an effective style similarity metric should align with human perceptions and definitions of style. Therefore, we have collected 508 styles, each handpicked by human artists and featuring four thematically distinct reference images, created using Midjourney's Sref mode \cite{midjourney_sref}. Midjourney's style reference feature allows users to guide the style or aesthetic of generated images by using external pictures or style seeds in their prompts. Figure. \ref{bench} shows some examples in our benchmark.

% 风格本质上是主观的，因此一个好的风格相似度评估指标应当与人类对风格的感知和定义保持一致。为此，我们在网站上收集了508个人类艺术家精选的风格，每个风格包含四张题材不同的参考图像， 由Midjourney 的Sref模式创建。 Midjourney 的样式参考允许用户通过在提示中使用外部图片或风格种子作为参考来指导 Midjourney 生成的图像的风格或美感。

\vspace{-0.4em}
\subsubsection{IP Bench}
\vspace{-0.2em}
Instance-level consistency is one of the primary tasks in customized generation. However, there is a lack of high-quality benchmarks for assessing character consistency. We have collected images of 299 IP characters and used advanced Flux models \cite{flux} and the IP-Adapter to create several variants of each character with different consistency weights.

% 角色一致性是客制化生成的最要任务之一，然而角色一致性的评估缺乏优质的Benchmark。 我们收集了300个IP角色形象，基于先进的Flux模型和IP-Adapter制作每个形象在不同一致性权重下的的若干变体。

\vspace{-0.9em}
\section{Experiment}
\vspace{-0.3em}
\subsection{Experimental Setting}
\vspace{-0.3em}
We implemented DiffSim based on Stable Diffusion 1.5, where the total time step \(T\) for the SD diffusion model is 1000. U-Net includes downsampling blocks, middle blocks, and upsampling blocks. We explore using features from both downsampling and upsampling blocks. We also conducted grid searches across different layers and denoising time steps for each task, reporting the best results among these choices; further details are provided in the supplementary materials. The default setting of DiffSim is using the self-attn layer, and the input image resolution of 512 $\times$ 512.

% 我们基于Stable diffusion 1.5 实现了DiffSim,  SD扩散模型总的时间步长T为1000。U-Net包含下采样块、中间块和上采样块。我们在下采样块和上采样块中提取特征。为了公平比较，我们还对每个任务进行网格搜索（不同层和不同的去噪时间步），并报告在这些选择中最好的结果， 更多层和去噪时间步的结果见补充材料。 当为单个图像提取特征时，我们使用一批随机噪声来获取平均特征图，默认的批量大小为8，输入图像分辨率设置为512*512。

\vspace{-0.4em}
\subsection{Baselines}
\vspace{-0.2em}
% 本研究选取了五种相似性评估方法作为基准，包括基于对比学习的SOTA方法CLIP，自监督学习的SOTA方法DINO v2，以及LPIPS、FFA和用于衡量风格相似性的Gram metric。所有对比的基线方法均基于预训练模型实现，确保能够公正地反映出基础模型在相似性评估任务上的性能。不包括那些涉及模型集成或在人类评估数据集上进行微调的方法。

We compares five similarity assessment methods as baselines, including CLIP \cite{clip}, DINO v2 \cite{dinov2}, as well as LPIPS \cite{lpips}, Foreground Feature Averaging (FFA) \cite{apple}, and the Gram metric \cite{gram} for style similarity assessment. All baseline methods used for comparison are implemented based on pre-trained models, ensuring a fair representation of the inherent performance differences in similarity metrics among foundational models. Methods involving fine-tuning on human-rated datasets \cite{dreamsim} are excluded from our comparison. Additionally, we demonstrate that an ensemble of CLIP, DINO v2 and DiffSim using a hard vote approach can improve performance.

% In addition, we also implement the ensembled method, which makes decision by comprehensively considering assessment from CLIP, DINO v2 and DiffSim.

\begin{table*}[ht]
\centering
\caption{Performance of different metrics across various benchmarks. Best results are highlighted in red and second-best in blue (excluding ensemble model). The ensemble model takes predictions from CLIP, DINO v2 and DiffSim and determines the final classification based on the majority rule. Results showing improvements over the original three methods in the ensemble model are highlighted in bold.}
\label{tab:performance}
\small 
\begin{tabular}{@{}c|cc|cc|c|cc@{}}
\toprule
\textbf{Model / Benchmark} & \multicolumn{2}{c|}{\textbf{Human-align Similarity}} & \multicolumn{2}{c|}{\textbf{Instance Similarity}} & \multicolumn{1}{c|}{\textbf{Low-level Similarity}} & \multicolumn{2}{c}{\textbf{Style Similarity}} \\ 
 & \textbf{NIGHTS} & \textbf{Dreambench++} & \textbf{CUTE} & \textbf{IP} & \textbf{TID2013} & \textbf{Sref}  & \textbf{InstantStyle bench} \\ \midrule
LPIPS\cite{lpips}                     & 71.13\%           & 62.33\%                  & 63.17\%          & 84.01\%          & 94.50\%           & 87.85\%                           & 93.15\%  \\
Gram\cite{gram}                      & -           & -                  & -          & -          & -           & 84.05\%                          & 88.30\%  \\
CLIP\cite{clip}                      & 82.26\%           & 70.54\%                  & 72.71\%          & \textcolor{blue}{91.70\%}          & 90.33\%           & 84.60\%                           & 82.90\%  \\
DINO v2\cite{dinov2}                   & 85.24\%           & \textcolor{red}{72.25\%}                  & \textcolor{red}{77.27\%}          & 85.35\%          & 91.50\%           & 87.20\%                           & 86.10\%  \\
FFA\cite{apple}                       & 77.78\%           & 65.21\%                  & 76.55\%          & 89.70\%          & 75.17\%           & 62.55\%                            & 60.65\%  \\ \midrule
CLIP AAS (ours)               & 82.18\%           & 67.23\%                  & 71.78\%          & 88.03\%          & \textcolor{red}{96.33\%}           & 94.45\%                            & \textcolor{blue}{97.80\%}  \\
DINO v2 AAS (ours)              & \textcolor{blue}{86.47\%}           & 70.01\%                  &  \textcolor{blue}{77.04\%}          & 90.38\%          & \textcolor{blue}{95.83\%}           & \textcolor{blue}{95.90\%}                        & 96.65\%  \\
DiffSim (ours)                   & \textcolor{red}{86.52\%}           & \textcolor{blue}{71.50\%}                  & 76.17\%          & \textcolor{red}{91.84\%}          & 94.17\%           & \textcolor{red}{97.40\%}                         & \textcolor{red}{99.05}\%  \\ 
\midrule
Ensemble                  & \textbf{89.43\%}          &       72.15\%           & \textbf{77.78\%}          & \textbf{94.92\%}          &     \textbf{95.00\%}      &               91.70\%            &  88.30\% \\
\bottomrule
\end{tabular}
\end{table*}

\begin{table*}[htbp]
    \centering
    \hfill
    \hspace*{-0.6cm}
    \begin{minipage}{0.4\textwidth}
        \centering
        \caption{Evaluation of different DiffSim architectures. The best results are highlighted in bold.}
        \label{tab:diffsim_setting}
        \setlength{\tabcolsep}{2pt}
        \footnotesize 
        \begin{tabular}{@{}lccc@{}}
        \toprule
        \textbf{Setting} & DiffSim-S SD1.5 & DiffSim-C SD1.5 & DiffSim-S SD-XL \\ \midrule
        NIGHTS      & \textbf{86.52\%} & 79.16\% & 78.05\% \\
        Dreambench++      & \textbf{71.50}\%  & 67.45\% & 63.93\% \\
        CUTE      & 72.06\%  & \textbf{76.17\%} & 69.94\% \\
        IP      & \textbf{92.04\%} & 77.06\% & 83.41\% \\
        TID2013      & \textbf{94.17}\%  & 94.00\% & 91.33\% \\
        Sref       & \textbf{97.40\%} & 94.70\% & 93.05\% \\
        InstantStyle      & \textbf{99.05}\% &  95.10\% & 96.55\% \\        
        \bottomrule
        \end{tabular}
    \end{minipage}%
    \hfill
    \hspace*{1.1cm}
    \begin{minipage}{0.33\textwidth}
        \centering
        \caption{Ablation study. CLIP and DINO v2 features (both AAS and original) are all extracted from the last layer. }
        \label{tab:ablation}
        \setlength{\tabcolsep}{4pt}
        \footnotesize 
        \begin{tabular}{@{}lccc@{}}
            \toprule
            \textbf{Benchmark} & Sref & IP & NIGHTS \\ \midrule
            DiffSim & \textbf{97.40}\% & \textbf{92.04}\% & \textbf{86.82}\% \\
            % DiffSim w/ MSE & 96.70\% & 92.11\% & 83.47\% \\
            Diffusion feature & 78.80\% & 62.47\% & 66.75\% \\
            % Diffusion feature (MSE) & x\% & x\% & x\% \\
            CLIP AAS & 71.15\% & 87.36\% & 80.54\% \\
            CLIP features & 66.50\% & 82.54\% & 75.84\% \\
            DINO v2 AAS & 78.50\% & 90.38\% & 86.41\% \\
            DINO v2 feature & 76.90\% & 87.56\% & 81.00\% \\
            \bottomrule
            \end{tabular}
    \end{minipage}%
    \hfill
    \begin{minipage}{0.22\textwidth}
        \centering
        \caption{Appearance consistency assessment in video. }
        \label{tab:tiktok}
        \setlength{\tabcolsep}{4pt}
        \footnotesize 
        \begin{tabular}{@{}lc@{}}
        \toprule
        \textbf{Metrics} & \textbf{Variance} $\downarrow$ \\ \midrule
         LPIPS &  3.565e-3\\
         CLIP & 1.007e-3 \\
         DINO v2 & 6.340e-3 \\
         FFA & 4.381e-3 \\
        % \midrule
        CLIP AAS & \textbf{4.397e-7} \\
        DINO v2 AAS & 1.825e-3\\
        DiffSim & 1.335e-3\\
        \bottomrule
        \end{tabular}
    \end{minipage}
    \hspace*{-0.5cm}
\end{table*}

\vspace{-0.4em}
\subsection{Benchmarks}
\vspace{-0.2em}
In addition to using the Sref bench and IP bench we proposed, we also employed 6 existing benchmarks.

\noindent \textbf{NIGHTS Perceptual Dataset.}
The NIGHTS\cite{dreamsim} (Novel Image Generations with Human-Tested Similarities) dataset comprises 20,019 image triplets, each containing a reference image and two distortions, evaluated for perceptual similarity by human observers. The dataset was generated using Stable Diffusion 2.1 with prompts from categories across prominent datasets like ImageNet and CIFAR. 

\noindent \textbf{TID2013.}
The TID2013\cite{tid2013} (Tampere Image Database 2013) dataset contains 3,000 test images derived from 25 reference images, each subjected to 24 types of distortions, producing 120 distorted images per reference. These distortions encompass a range of realistic noise and artifacts intended to evaluate image quality assessment metrics. 

\noindent \textbf{CUTE Benchmark.}
CUTE\cite{apple} includes 18,000 images of 180 objects across 50 categories under varied conditions to test intrinsic object-centric similarity. It features objects in diverse poses and settings, providing a robust resource for benchmarking similarity metrics in varying visual contexts.

\noindent \textbf{Dreambench++.}
Dreambench++\cite{dreambench++} serves as a human-aligned, automated benchmark for personalized image generation model evaluation, focusing on concept preservation and prompt adherence. It utilizes multimodal GPT models aligned with human preferences, with a refined rating system for efficient method comparison.

\noindent \textbf{InstantStyle Benchmark.}
InstantStyle\cite{instantstyle} is an advanced method for style customization. We have organized 30 styles on the InstantStyle project homepage, each including 5 images, to serve as a supplementary benchmark for style evaluation.

\noindent \textbf{TikTok Dataset.} The TikTok\cite{tictok} dataset includes 300 dance videos (10-15 seconds), which we use to assess video appearance consistency in this paper.

% TikTok数据集包括300个 dance videos(10-15 seconds)，本文中我们用于 Appearance Consistency Assessment in Video
% InstantStyle是先进的风格客制化方法。我们在InstantStyle的项目主页整理了30个风格，每个包括5张图片， 作为风格评估的补充Benchmark。

\begin{figure*}[htp]
    \centering
    \includegraphics[width=1.0\linewidth]{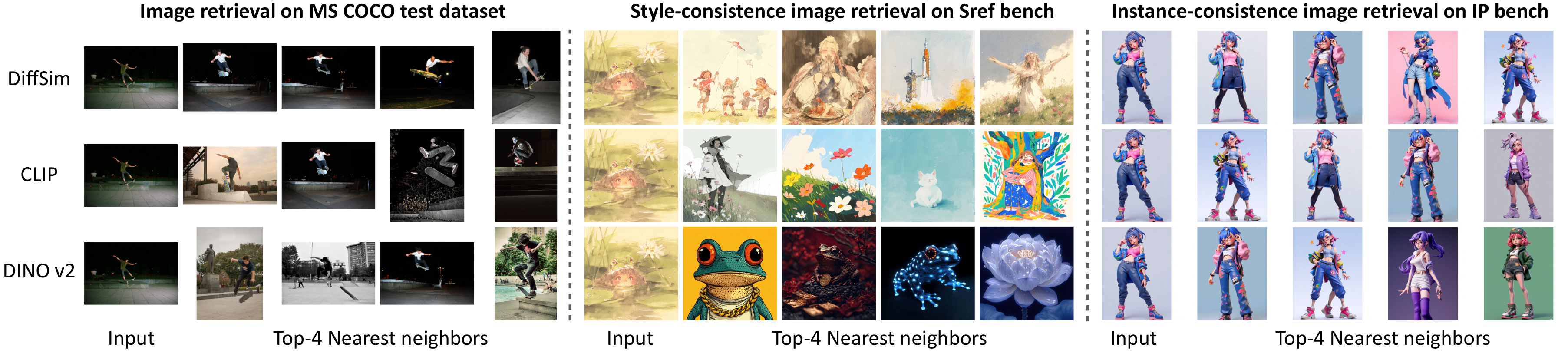}
    \caption{Some retrieval examples using DiffSim, CLIP, and DINO v2. The left, middel and right column displays retrieval results from the Sref benchmark, MS COCO Test dataset and the IP benchmark respectively.}
    \label{fig:retrieval}
\end{figure*}

\begin{figure}[htp]
    \includegraphics[width=1.0\linewidth]{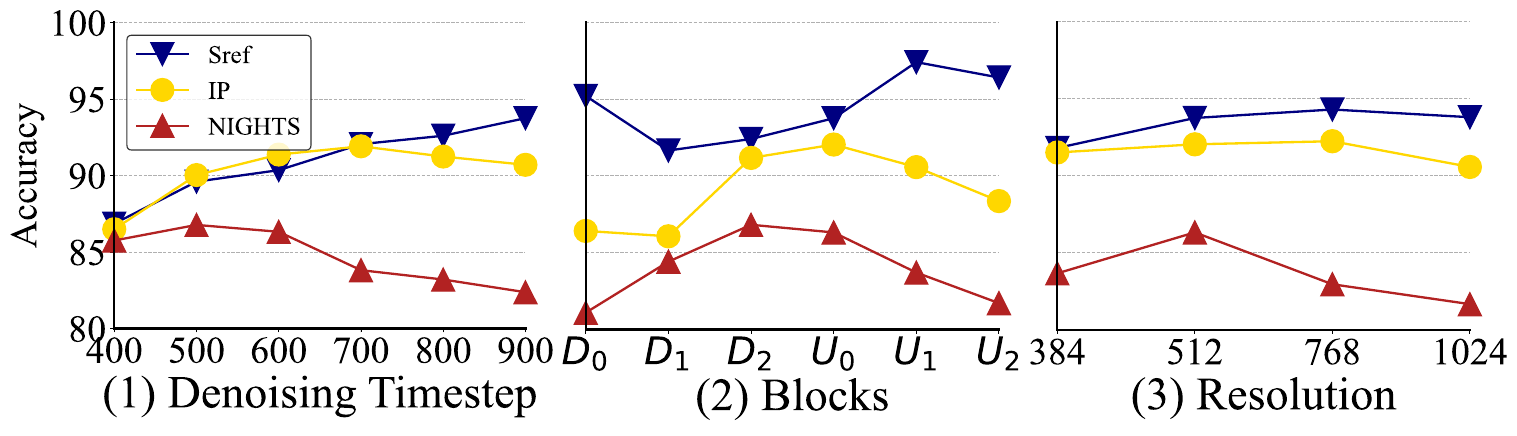}
    \caption{Evaluation of different benchmarks across different timesteps, blocks and resolutions. (1) All experiments across timesteps are conducted on the $\text{U}_0$ block of SD 1.5. (2) The results of different blocks of Sref, IP, and NIGHTS experiments are completed with fixed $t$ of 900, 750, and 600, respectively. (3) The results of different resolutions are all based on the best settings of timesteps and blocks.}
    \label{fig:plots}
\end{figure}

\vspace{-0.5em}
\subsection{Quantitative Evaluation}
\vspace{-0.3em}
\label{sec:quant}
% 定量评估 
% 我们使用与之前的方法（Apple， DreamSim）相同的评估模式。 分别计算一张reference image和2张候选图像的相似性得分，得分高的图片作为DiffSim的选择，并统计分类正确的准确率。各个度量在所有基准测试中的准确率结果显示在表 ~\ref{tab:performance} 中。不同基准测试的候选图像选择方式稍有不同，详细信息请参阅补充材料。

We employ the same evaluation protocol as previous methods \cite{dreamsim}. Similarity scores are computed between a reference image and two candidate images. Image with the higher score is selected as the choice of current evaluated model, and we tally the accuracy of correctly classified selections for each metric model. The accuracy results of each metric across all benchmarks are displayed in the Table ~\ref{tab:performance}. The selection of candidate images varies slightly for different benchmarks; for further details, please refer to the supplementary materials.

\noindent \textbf{Human Perceptual Consistency Assessment.} To evaluate the overall similarity and consistency between DiffSim and human judgments, we conducted experiments on the NIGHTS and Dreambench++ benchmarks. As seen in Table ~\ref{tab:performance}, DiffSim achieved the best performance on the Nights dataset, while on the Dreambench dataset, its performance is comparable to that of the top-performing DINO v2.

%为了评估整体相似度DiffSim和人类判断的一致性，我们在NIGHTS，Dreambench++ 进行实验。 Table ~\ref{tab:performance 可以看出，DiffSim的结果比CLIP和DINO v2都要好，和DreamSim 方法媲美，然而DreamSim 是采用CLIP 和DINO Encoder模型集成的结果，而且在NIGHTS数据集上微调了。值得一提的是， 我们提出的CLIP Cross-Sim 和 DINO v2 Cross-Sim 和CLIP和DNIO v2相比也有不同程度的提升。

\noindent \textbf{Instance Similarity Assessment.} To compare the accuracy of instance-level metrics, experiments were conducted on the CUTE and IP benchmarks. On the CUTE dataset, DINO v2 and FFA take the lead, but their performance does not surpass DiffSim by a large margin. On the IP benchmark, DiffSim leads with a significant margin. It shows that DiffSim is capable to capture instance level similarity.

%为了比较实例级的度量的准确性， 我们在CUTE benchmark和IP benchmark进行实验。在CUTE数据集上，DreamSim和FFA处于领先地位， Dim-Sim紧随其后，比CLIP等其他指标明显更好。在IP benchmark上， DiffSim 结果以绝对优势领先, 基于扩散模型实现的DiffSim在合成数据评估上优势显著。

\noindent \textbf{Low-level Similarity Assessment.} To evaluate the accuracy of various methods, experiments were also conducted on the TID2013 dataset. The DiffSim method showed considerable versatility, achieving excellent performance in low-level similarity assessment merely by modifying the denoising time step, far surpassing CLIP and DINO v2, and comparable to LPIPS.
% 为了评估各个方法在的准确性，我们还在TID2013上进行了实验，DiffSim方法展现出相当的通用性，仅通过修改去噪time step， 就在low-level similarity评估方面取得很好的表现，远远超出CLIP和DINO v2，和LPIPS媲美。值得一提的是， 我们提出的CLIP Cross-Sim 和 DINO v2 Cross-Sim 和CLIP和DNIO v2相比有显著的提升，进一步证明我们方法的贡献。

\noindent \textbf{Style Similarity Assessment.} To assess the accuracy of style similarity metrics, experiments were conducted on the Sref bench and InstantStyle bench. Across these two benchmarks, DiffSim achieved the best results, demonstrating its strong ability in evaluating stylish similarity.
% 为了评估风格相似性度量的准确性， 我们在Sref bench, Instant-Style bench, Fruit-SALAD上进行实验。在3个Bench上，Sref bench上， DiffSim取得了最好的成绩。Fruit-SALAD 只包含10中水果和10种容易区分的风格，所有的指标均取得了很好的结果.

\noindent \textbf{Appearance Consistency Assessment in Video.}
Temporal appearance consistency is crucial for assessing video generation and video-to-image models. An ideal metric should be unaffected by changes in object position and layout, ensuring stable similarity scores across frames of the same subject, regardless of movement. We evaluate on the TikTok dataset by measuring variance of different similarity scores between the first frame and other frames. As shown in Table ~\ref{tab:tiktok}, our CLIP AAS outperforms others, with CLIP and DiffSim yielding similar results.
% Temporal appearance consistency is an important aspect of evaluating video generation models and video-to-image models. An ideal temporal appearance consistency metric should be insensitive to changes in object position and layout: in real videos, the similarity scores between different frames of the same subject should not fluctuate significantly with the movement of the subject. Therefore we conduct experiments on the Tiktok dataset, where we measure the variance of each metrics. As indicated in Table ~\ref{tab:tiktok}, our CLIP AAS achieves the best results, with CLIP and DiffSim following with similar performance.
% 时序外观一致性。 时序外观一致性是评估视频生成模型和图生视频模型的重要方面，理想的时序外观一致性评估指标应对物体的位置和layout变化不敏感：在真实视频上，视频中同一个主体的不同帧的相似性数值不应随主体运动产生明显波动。当主体发生外观变化，评价指标应能准确反映外观变化程度。我们在TikToc数据集上进行实验，通过逐帧计算所有帧和首帧的相似性，并统计均值和方差。

\noindent \textbf{CLIP AAS and DINO v2 AAS Assessment} As indicated in the Table ~\ref{tab:performance} and ~\ref{tab:tiktok}, CLIP AAS and DINO v2 AAS show largely comparable results to CLIP and DINO v2, with some improvements on specific benchmarks. This further demonstrates the effectiveness of the proposed AAS and its scalability across various attention-based frameworks.

\noindent \textbf{Ensemble Model} We show that by ensembling CLIP, DINO v2 and DiffSim models through hard vote, the performance has shown an improvement over three orginal methods across several benchmarks. This demonstrates that DiffSim can be used to compensate for the limitations of the CLIP and DINO v2 models.

\vspace{-0.4em}
\subsection{Performance Analysis}
\vspace{-0.3em}

In this section, we select three representative benchmarks to demonstrate the key factors affecting DiffSim's performance in three similarity evaluation tasks.

\noindent \textbf{Denoising Timestep.} Our observations reveal that diffusion models manifest distinct characteristics at different denoising timesteps. As shown in Figure ~\ref{fig:plots}, higher $t$ values (900) cater to style assessments with rich low-level features, while lower $t$ values (500) excel in instance-level similarity tasks on the NIGHTS dataset.

\noindent \textbf{Different Attention Blocks.} Results from IP bench and NIGHTS dataset show that layers near the middle of the Unet (Down sample layer 2, Up sample layer 0) focus more on instance level, while those closer to the ends of the network (Down sample layer 0, Up sample layer 1) excel on the Sref bench. We believe that shallow features are more advantageous for assessing style similarity.

\noindent \textbf{Resolution.} We test resolutions from 384 to 1024, with no notable impact across most datasets. However, increasing the resolution from 512 to 768 slightly improves performance on the IP and Sref benches due to their original higher resolution, unlike the NIGHTS dataset, which did not benefit from higher inference resolutions. We standardize the resolution at 512x512 to align with Stable Diffusion 1.5’s training settings.

\noindent \textbf{Different DiffSim Architectures.} Table ~\ref{tab:diffsim_setting} demonstrates the impact of different DiffSim architectures. Overall, DiffSim-S SD1.5 shows superior performance. For instance-level similarity assessments on the CUTE dataset, DiffSim-C SD1.5 performs slightly better.

%本节，我们选择有代表性的3个bench，来呈现人类一致性评估、风格相似度评估、实例相似度评估任务中， 影响DifSim性能的关键因素。
% 1.去噪时间步。我们发现扩散模型在不同去噪时间步的特征具有不同的属性。如表2所示，Sref bench 上，较大的t值(900)得到的特征包含更多低层次信息，适合于风格评估。NIGHTSS dataset上，而较小的t值(400)对应深层的语义信息，在human-align similarity评估任务中表现更为出色。IP bench上，最好的t值对应的是700。
% 2. 如图4（2)所示， 在IP bench 和 NIGHTS dataset上， 更靠近Unet中间的层（Down sample layer2，Up sample layer0）表现更好，IP bench 上更靠近两端的层（Down sample layer0，Up sample layer1）表现更好， 我们认为浅层特征对风格相似性的判断更有利。
% 3. 分辨率。此外，我们也探讨了分辨率对相似度评估结果的影响，实验设置了多种分辨率：384、512、768和1024，如图4（3)所示，当分辨率从512提升到768，IP bench 和Sref bench上，分辨率的提升带来微小的性能提升，这是因为这两个Bench的原始图片是1024分辨率，resize到768仍可以提供更多有效信息，而NIGHTS dataset原始图片是512分辨率，增加推理的分辨率没有提升。综上，我们选择512x512的分辨率作为默认分辨率，与Stable Diffusion 1.5的训练分辨率保持一致。
% 4. 不同架构。表2展示了DiffSim不同架构的影响，在SD1.5 自注意力层实现的DiffSim整体性能领先，在用于实例相似度评估的CUTE数据集上， 在交叉注意力层实现的DiffSim-IPA SD1.5 稍好一些。

\vspace{-0.5em}
\subsection{Ablation Study}
\vspace{-0.5em}
% 本节，我们展示消融实验结果，来证明本文提出的关键设计，AAS的有效性。 作为对比， 我们控制层和time step相同， 直接计算L_{A} 和 L_{B}的cos similarity 和 MSE， 此外，我们计算AAS的MSE。如，表X所示，当不使用特征对齐，直接计算MSE和cos similarity效果有显著下降。6行
% DiffSim, Dift cos, Dift mse, clip specific layer feats, clip AAS, dino specific layer feats, dino AAS

% In this section, we present the results of ablation studies to demonstrate the effectiveness of the key design introduced in this paper, the Aligned Attention Score (AAS). For comparison, we control for the same layers and time steps, directly computing the cosine similarity and Mean Squared Error (MSE) between \(L_{A}\) and \(L_{B}\). Additionally, we calculate the MSE of AAS. As shown in Table ~\ref{tab:ablation}, the results significantly deteriorate when feature alignment is not used and cosine similarity and MSE are calculated directly. 

In this section, we present the results of ablation studies to demonstrate the effectiveness of the key design introduced in this paper, the Aligned Attention Score (AAS). For comparison, we fix the layer and timestep setting, directly calculating the cosine similarity of latent diffusion model features, CLIP features and DINO v2 features $L_A$, $L_B$. As shown in Table ~\ref{tab:ablation}, the results significantly deteriorate when AAS feature alignment is not used.

% \begin{table}[ht]
% \centering
% \caption{Ablation Study.}
% \label{tab:ablation}
% \setlength{\tabcolsep}{4pt}
% \footnotesize 
% \begin{tabular}{@{}lccc@{}}
% \toprule
% Benchmark & Sref & IP & NIGHTS \\ \midrule
% DiffSim & \textbf{97.40}\% & \textbf{92.04}\% & \textbf{86.82}\% \\
% DiffSim w/ MSE & 96.70\% & 92.11\% & 83.47\% \\
% Diffusion feature & 78.80\% & 62.47\% & 66.75\% \\
% % Diffusion feature (MSE) & x\% & x\% & x\% \\
% CLIP features & 66.50\% & 82.54\% & 75.84\% \\
% DINO v2 feature & 76.90\% & 87.56\% & 81.00\% \\
% \bottomrule
% \end{tabular}
% \end{table}

\vspace{-0.5em}
\subsection{Image Retrieval}
\vspace{-0.5em}
To further analyze DiffSim's performance, we conduct image retrieval experiments on the Sref bench, IP bench, and MS COCO Test dataset, with Figure ~\ref{fig:retrieval} showcasing the top 4 nearest neighbors. During style consistency retrieval on the Sref bench, our results are quite precise, while baseline methods are often influenced by image category attributes. On the COCO Test dataset, our retrieval results focus not only on semantics but also on background similarity, which is better than the baseline methods. On the IP bench, our results are more precise. Please refer to the supplementary materials for more retrieval examples.

% 为了进一步分析DiffSim的表现，我们在Sref bench, IP bench 和CoCo Val dataset上做了图像检索实验, 图4展示了top5 nearest neighbors。在Sref bench进行风格一致性的检索时，我们的结果相当精准，而baseline方法会常常受图像类别属性干扰。在CoCo Val dataset数据集上，我们的检索结果不止关注语义，背景的相似性比baseline方法更好。在 IP bench 上， 我们的结果更加精准。

\vspace{-1em}
\begin{figure}[htp]
    \includegraphics[width=0.9\linewidth]{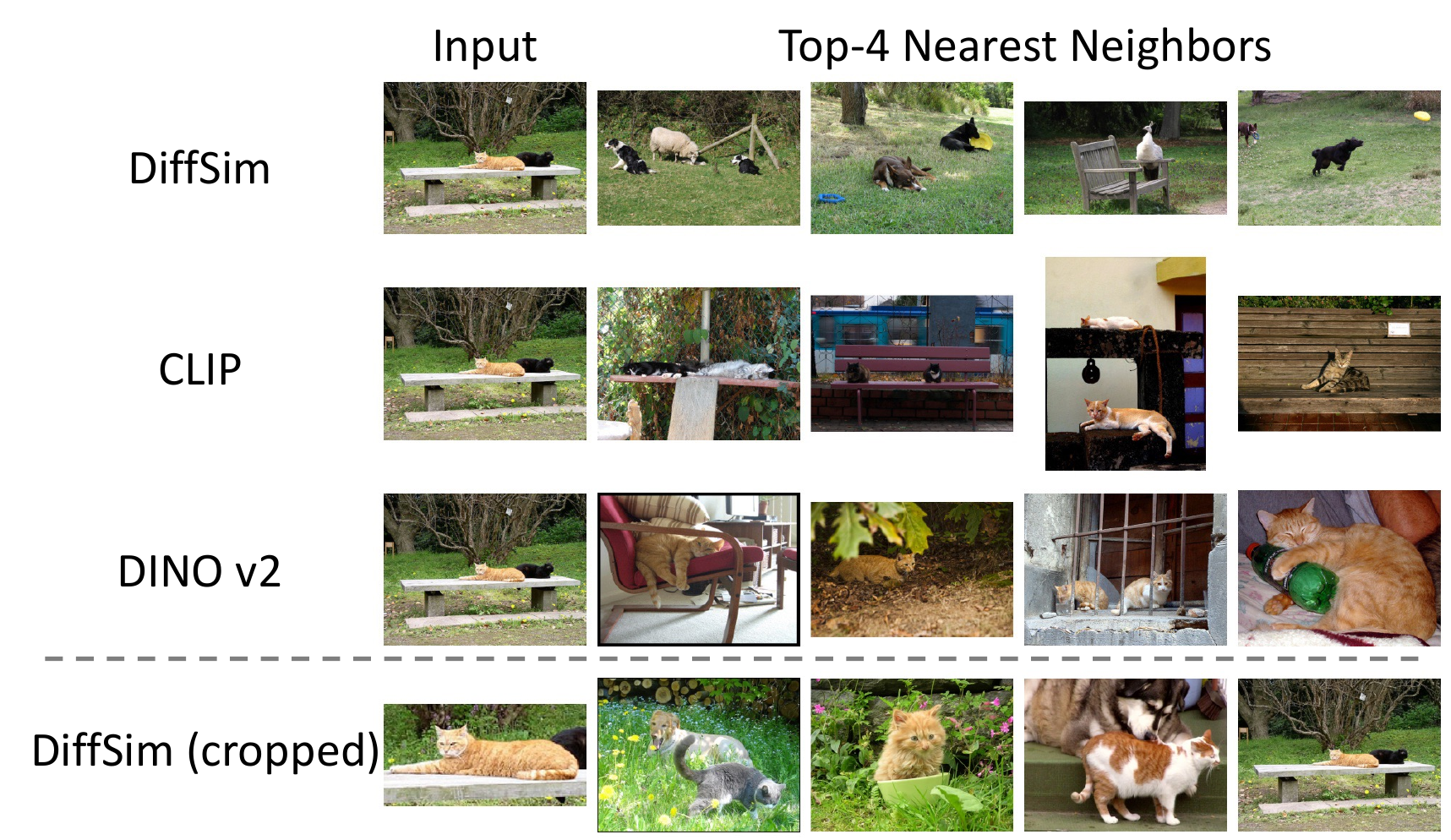}
    \vspace{-0.5em}
    \caption{Failure case.}
    \label{fig:fail}
\end{figure}

\vspace{-1em}
\subsection{Limitation and Failure Case}
% DiffSim occasionally overemphasizes background features, neglecting key subject details, especially in images with smaller subjects. As Figure \ref{fig:fail} shows, DiffSim uses an image of cats to retrieve some dogs in a similar background. CLIP and DINO show better performance in these scenarios.

% DiffSim 有时会过度强调背景特征，而忽略了关键主体的细节，尤其是在主体较小的图像中。正如图 \ref{fig:fail} 所示，DiffSim 使用一张包含猫的图像检索到了具有相似背景的狗的图像。对主体进行crop可以缓解这一问题。

DiffSim occasionally overemphasizes background features, neglecting key subject details, especially in images with smaller subjects. As shown in Figure \ref{fig:fail}, DiffSim retrieves images of dogs with similar backgrounds using an image of cats. Cropping the subject can mitigate this issue.

\vspace{-0.5em}
\section{Conclusion}
\vspace{-0.5em}
This paper introduces DiffSim, marking the first application of diffusion models in image similarity assessment. DiffSim calculates the Aligned Attention Score in the self-attention or cross-attention layers of Stable Diffusion, effectively evaluating visual similarity and demonstrating high alignment with human perception across various tasks. By exploring the characteristics of different layers and denoising steps within diffusion models, DiffSim achieves accurate assessments of human-aligned, instance-level, and style similarity. Extensive experiments on multiple benchmark datasets confirm the effectiveness of DiffSim, providing an efficient tool for evaluating visual consistency.

% 本文提出了DiffSim方法，首次将扩散模型用于图像相似性评估。DiffSim在Stable diffusion层的自注意力或交叉注意力层中计算Aligned Attention Score（AAS），有效评估视觉相似性，在多种任务上展现出与人类感知高度一致的优势。通过探索扩散模型不同层次与去噪步数的特性，DiffSim实现了精确的human-align、实例级别和风格相似性评估。我们在多个基准数据集上进行了详尽实验，验证了DiffSim的优越性，为生成模型的视觉一致性评估提供了一种高效工具。

{
    \small
    \bibliographystyle{ieeenat_fullname}
    \bibliography{main}

\begin{thebibliography}{62}
\providecommand{\natexlab}[1]{#1}
\providecommand{\url}[1]{\texttt{#1}}
\expandafter\ifx\csname urlstyle\endcsname\relax
  \providecommand{\doi}[1]{doi: #1}\else
  \providecommand{\doi}{doi: \begingroup \urlstyle{rm}\Url}\fi

\bibitem[mid()]{midjourney_sref}
Midjourney sref: Prompt library and examples.
\newblock \url{https://midjourneysref.com/discover?page=1}.
\newblock Accessed: 2024-11-21.

\bibitem[AI(2024)]{flux}
Flux.1 AI.
\newblock Flux.1 ai, 2024.

\bibitem[Alaluf et~al.(2024)Alaluf, Garibi, Patashnik, Averbuch-Elor, and Cohen-Or]{appearance}
Yuval Alaluf, Daniel Garibi, Or Patashnik, Hadar Averbuch-Elor, and Daniel Cohen-Or.
\newblock Cross-image attention for zero-shot appearance transfer.
\newblock In \emph{ACM SIGGRAPH 2024 Conference Papers}, pages 1--12, 2024.

\bibitem[Blattmann et~al.(2023)Blattmann, Dockhorn, Kulal, Mendelevitch, Kilian, Lorenz, Levi, English, Voleti, Letts, et~al.]{svd}
Andreas Blattmann, Tim Dockhorn, Sumith Kulal, Daniel Mendelevitch, Maciej Kilian, Dominik Lorenz, Yam Levi, Zion English, Vikram Voleti, Adam Letts, et~al.
\newblock Stable video diffusion: Scaling latent video diffusion models to large datasets.
\newblock \emph{arXiv preprint arXiv:2311.15127}, 2023.

\bibitem[Brooks et~al.(2023)Brooks, Holynski, and Efros]{instructpix2pix}
Tim Brooks, Aleksander Holynski, and Alexei~A Efros.
\newblock Instructpix2pix: Learning to follow image editing instructions.
\newblock In \emph{Proceedings of the IEEE/CVF Conference on Computer Vision and Pattern Recognition}, pages 18392--18402, 2023.

\bibitem[Caron et~al.(2021)Caron, Touvron, Misra, J’egou, Mairal, Bojanowski, and Joulin]{dinov1}
Mathilde Caron, Hugo Touvron, Ishan Misra, Herv{\'e} J’egou, Julien Mairal, Piotr Bojanowski, and Armand Joulin.
\newblock Emerging properties in self-supervised vision transformers. 2021 ieee.
\newblock In \emph{CVF International Conference on Computer Vision (ICCV)}, 2021.

\bibitem[Chen et~al.(2023{\natexlab{a}})Chen, Dong, Wang, Yang, Duan, Su, and Zhu]{chen2023robust}
Huanran Chen, Yinpeng Dong, Zhengyi Wang, Xiao Yang, Chengqi Duan, Hang Su, and Jun Zhu.
\newblock Robust classification via a single diffusion model.
\newblock \emph{arXiv preprint arXiv:2305.15241}, 2023{\natexlab{a}}.

\bibitem[Chen et~al.(2023{\natexlab{b}})Chen, Liu, Chen, Feng, Liu, Shen, and Zhao]{livephoto}
Xi Chen, Zhiheng Liu, Mengting Chen, Yutong Feng, Yu Liu, Yujun Shen, and Hengshuang Zhao.
\newblock Livephoto: Real image animation with text-guided motion control.
\newblock \emph{arXiv preprint arXiv:2312.02928}, 2023{\natexlab{b}}.

\bibitem[Chen et~al.(2024)Chen, Cao, Chen, Li, and Ma]{echomimic}
Zhiyuan Chen, Jiajiong Cao, Zhiquan Chen, Yuming Li, and Chenguang Ma.
\newblock Echomimic: Lifelike audio-driven portrait animations through editable landmark conditions.
\newblock \emph{arXiv preprint arXiv:2407.08136}, 2024.

\bibitem[Deng et~al.(2019)Deng, Guo, Xue, and Zafeiriou]{arcface}
Jiankang Deng, Jia Guo, Niannan Xue, and Stefanos Zafeiriou.
\newblock Arcface: Additive angular margin loss for deep face recognition.
\newblock In \emph{Proceedings of the IEEE/CVF conference on computer vision and pattern recognition}, pages 4690--4699, 2019.

\bibitem[El~Banani et~al.(2024)El~Banani, Raj, Maninis, Kar, Li, Rubinstein, Sun, Guibas, Johnson, and Jampani]{probing}
Mohamed El~Banani, Amit Raj, Kevis-Kokitsi Maninis, Abhishek Kar, Yuanzhen Li, Michael Rubinstein, Deqing Sun, Leonidas Guibas, Justin Johnson, and Varun Jampani.
\newblock Probing the 3d awareness of visual foundation models.
\newblock In \emph{Proceedings of the IEEE/CVF Conference on Computer Vision and Pattern Recognition}, pages 21795--21806, 2024.

\bibitem[Fu et~al.(2023)Fu, Tamir, Sundaram, Chai, Zhang, Dekel, and Isola]{dreamsim}
Stephanie Fu, Netanel Tamir, Shobhita Sundaram, Lucy Chai, Richard Zhang, Tali Dekel, and Phillip Isola.
\newblock Dreamsim: Learning new dimensions of human visual similarity using synthetic data.
\newblock \emph{arXiv preprint arXiv:2306.09344}, 2023.

\bibitem[Gatys(2015)]{gram}
Leon~A Gatys.
\newblock A neural algorithm of artistic style.
\newblock \emph{arXiv preprint arXiv:1508.06576}, 2015.

\bibitem[Gatys et~al.(2016)Gatys, Ecker, and Bethge]{gatys2016}
Leon~A. Gatys, Alexander~S. Ecker, and Matthias Bethge.
\newblock Image style transfer using convolutional neural networks.
\newblock In \emph{2016 IEEE Conference on Computer Vision and Pattern Recognition (CVPR)}, pages 2414--2423, 2016.

\bibitem[Guo et~al.(2024)Guo, Zheng, Hou, Gao, Deng, Wan, Zhang, Liu, Hu, Zha, et~al.]{i2vadapter}
Xun Guo, Mingwu Zheng, Liang Hou, Yuan Gao, Yufan Deng, Pengfei Wan, Di Zhang, Yufan Liu, Weiming Hu, Zhengjun Zha, et~al.
\newblock I2v-adapter: A general image-to-video adapter for diffusion models.
\newblock In \emph{ACM SIGGRAPH 2024 Conference Papers}, pages 1--12, 2024.

\bibitem[Guo et~al.(2023)Guo, Yang, Rao, Wang, Qiao, Lin, and Dai]{animatediff}
Yuwei Guo, Ceyuan Yang, Anyi Rao, Yaohui Wang, Yu Qiao, Dahua Lin, and Bo Dai.
\newblock Animatediff: Animate your personalized text-to-image diffusion models without specific tuning.
\newblock \emph{arXiv preprint arXiv:2307.04725}, 2023.

\bibitem[He et~al.(2024)He, Liu, Qian, Wang, Hu, Cao, Yan, Zhou, and Zhang]{idanimator}
Xuanhua He, Quande Liu, Shengju Qian, Xin Wang, Tao Hu, Ke Cao, Keyu Yan, Man Zhou, and Jie Zhang.
\newblock Id-animator: Zero-shot identity-preserving human video generation.
\newblock \emph{arXiv preprint arXiv:2404.15275}, 2024.

\bibitem[Hebart et~al.(2019)Hebart, Dickter, Kidder, Kwok, Corriveau, Van~Wicklin, and Baker]{things}
Martin~N Hebart, Adam~H Dickter, Alexis Kidder, Wan~Y Kwok, Anna Corriveau, Caitlin Van~Wicklin, and Chris~I Baker.
\newblock Things: A database of 1,854 object concepts and more than 26,000 naturalistic object images.
\newblock \emph{PloS one}, 14\penalty0 (10):\penalty0 e0223792, 2019.

\bibitem[Hertz et~al.(2022)Hertz, Mokady, Tenenbaum, Aberman, Pritch, and Cohen-Or]{p2p}
Amir Hertz, Ron Mokady, Jay Tenenbaum, Kfir Aberman, Yael Pritch, and Daniel Cohen-Or.
\newblock Prompt-to-prompt image editing with cross attention control.
\newblock \emph{arXiv preprint arXiv:2208.01626}, 2022.

\bibitem[Ho et~al.(2020)Ho, Jain, and Abbeel]{ddpm}
Jonathan Ho, Ajay Jain, and Pieter Abbeel.
\newblock Denoising diffusion probabilistic models.
\newblock \emph{Advances in neural information processing systems}, 33:\penalty0 6840--6851, 2020.

\bibitem[Hu et~al.(2021)Hu, Shen, Wallis, Allen-Zhu, Li, Wang, Wang, and Chen]{hu2021loralowrankadaptationlarge}
Edward~J. Hu, Yelong Shen, Phillip Wallis, Zeyuan Allen-Zhu, Yuanzhi Li, Shean Wang, Lu Wang, and Weizhu Chen.
\newblock Lora: Low-rank adaptation of large language models, 2021.

\bibitem[Hu et~al.(2022)Hu, Shen, Wallis, Allen-Zhu, Li, Wang, Wang, and Chen]{lora}
Edward~J Hu, Yelong Shen, Phillip Wallis, Zeyuan Allen-Zhu, Yuanzhi Li, Shean Wang, Lu Wang, and Weizhu Chen.
\newblock Lo{RA}: Low-rank adaptation of large language models.
\newblock In \emph{International Conference on Learning Representations}, 2022.

\bibitem[Hu(2024)]{animateanyone}
Li Hu.
\newblock Animate anyone: Consistent and controllable image-to-video synthesis for character animation.
\newblock In \emph{Proceedings of the IEEE/CVF Conference on Computer Vision and Pattern Recognition}, pages 8153--8163, 2024.

\bibitem[Hughes et~al.(2011)Hughes, Graham, Jacobsen, and Rockmore]{hughes2011}
James~M. Hughes, Daniel~J. Graham, C.~Robert Jacobsen, and Daniel~N. Rockmore.
\newblock Comparing higher-order spatial statistics and perceptual judgements in the stylometric analysis of art.
\newblock In \emph{2011 19th European Signal Processing Conference}, pages 1244--1248, 2011.

\bibitem[Jafarian and Park(2022)]{tictok}
Yasamin Jafarian and Hyun~Soo Park.
\newblock Self-supervised 3d representation learning of dressed humans from social media videos.
\newblock \emph{IEEE Transactions on Pattern Analysis and Machine Intelligence}, 45\penalty0 (7):\penalty0 8969--8983, 2022.

\bibitem[Kotar et~al.(2023)Kotar, Tian, Yu, Yamins, and Wu]{apple}
Klemen Kotar, Stephen Tian, Hong-Xing Yu, Dan Yamins, and Jiajun Wu.
\newblock Are these the same apple? comparing images based on object intrinsics.
\newblock \emph{Advances in Neural Information Processing Systems}, 36:\penalty0 40853--40871, 2023.

\bibitem[Krizhevsky et~al.(2012)Krizhevsky, Sutskever, and Hinton]{alexnet}
Alex Krizhevsky, Ilya Sutskever, and Geoffrey~E Hinton.
\newblock Imagenet classification with deep convolutional neural networks.
\newblock \emph{Advances in neural information processing systems}, 25, 2012.

\bibitem[Kumari et~al.(2023)Kumari, Zhang, Zhang, Shechtman, and Zhu]{customdiffusion}
Nupur Kumari, Bingliang Zhang, Richard Zhang, Eli Shechtman, and Jun-Yan Zhu.
\newblock Multi-concept customization of text-to-image diffusion.
\newblock In \emph{Proceedings of the IEEE/CVF Conference on Computer Vision and Pattern Recognition}, pages 1931--1941, 2023.

\bibitem[Luan et~al.(2017)Luan, Paris, Shechtman, and Bala]{luan2017deepphotostyletransfer}
Fujun Luan, Sylvain Paris, Eli Shechtman, and Kavita Bala.
\newblock Deep photo style transfer, 2017.

\bibitem[Mou et~al.(2023)Mou, Wang, Xie, Zhang, Qi, Shan, and Qie]{t2i}
Chong Mou, Xintao Wang, Liangbin Xie, Jian Zhang, Zhongang Qi, Ying Shan, and Xiaohu Qie.
\newblock T2i-adapter: Learning adapters to dig out more controllable ability for text-to-image diffusion models.
\newblock \emph{arXiv preprint arXiv:2302.08453}, 2023.

\bibitem[Muttenthaler et~al.(2022)Muttenthaler, Dippel, Linhardt, Vandermeulen, and Kornblith]{humanalign}
Lukas Muttenthaler, Jonas Dippel, Lorenz Linhardt, Robert~A Vandermeulen, and Simon Kornblith.
\newblock Human alignment of neural network representations.
\newblock \emph{arXiv preprint arXiv:2211.01201}, 2022.

\bibitem[Oquab et~al.(2023)Oquab, Darcet, Moutakanni, Vo, Szafraniec, Khalidov, Fernandez, Haziza, Massa, El-Nouby, et~al.]{dinov2}
Maxime Oquab, Timoth{\'e}e Darcet, Th{\'e}o Moutakanni, Huy Vo, Marc Szafraniec, Vasil Khalidov, Pierre Fernandez, Daniel Haziza, Francisco Massa, Alaaeldin El-Nouby, et~al.
\newblock Dinov2: Learning robust visual features without supervision.
\newblock \emph{arXiv preprint arXiv:2304.07193}, 2023.

\bibitem[Peebles and Xie(2023)]{dit}
William Peebles and Saining Xie.
\newblock Scalable diffusion models with transformers.
\newblock In \emph{Proceedings of the IEEE/CVF International Conference on Computer Vision}, pages 4195--4205, 2023.

\bibitem[Peng et~al.(2024)Peng, Cui, Tang, Qi, Dong, Bai, Han, Ge, Zhang, and Xia]{dreambench++}
Yuang Peng, Yuxin Cui, Haomiao Tang, Zekun Qi, Runpei Dong, Jing Bai, Chunrui Han, Zheng Ge, Xiangyu Zhang, and Shu-Tao Xia.
\newblock Dreambench++: A human-aligned benchmark for personalized image generation.
\newblock \emph{arXiv preprint arXiv:2406.16855}, 2024.

\bibitem[Ponomarenko et~al.(2013)Ponomarenko, Ieremeiev, Lukin, Jin, Egiazarian, Astola, Vozel, Chehdi, Carli, Battisti, et~al.]{tid2013}
Nikolay Ponomarenko, Oleg Ieremeiev, Vladimir Lukin, Lina Jin, Karen Egiazarian, Jaakko Astola, Benoit Vozel, Kacem Chehdi, Marco Carli, Federica Battisti, et~al.
\newblock A new color image database tid2013: Innovations and results.
\newblock In \emph{Advanced Concepts for Intelligent Vision Systems: 15th International Conference, ACIVS 2013, Pozna{\'n}, Poland, October 28-31, 2013. Proceedings 15}, pages 402--413. Springer, 2013.

\bibitem[Prashnani et~al.(2018)Prashnani, Cai, Mostofi, and Sen]{pieapp}
Ekta Prashnani, Hong Cai, Yasamin Mostofi, and Pradeep Sen.
\newblock Pieapp: Perceptual image-error assessment through pairwise preference.
\newblock In \emph{Proceedings of the IEEE Conference on Computer Vision and Pattern Recognition}, pages 1808--1817, 2018.

\bibitem[Radford et~al.(2021)Radford, Kim, Hallacy, Ramesh, Goh, Agarwal, Sastry, Askell, Mishkin, Clark, Krueger, and Sutskever]{clip}
Alec Radford, Jong~Wook Kim, Chris Hallacy, Aditya Ramesh, Gabriel Goh, Sandhini Agarwal, Girish Sastry, Amanda Askell, Pamela Mishkin, Jack Clark, Gretchen Krueger, and Ilya Sutskever.
\newblock Learning transferable visual models from natural language supervision.
\newblock In \emph{Proceedings of the 38th International Conference on Machine Learning}, pages 8748--8763. PMLR, 2021.

\bibitem[Rombach et~al.(2022)Rombach, Blattmann, Lorenz, Esser, and Ommer]{rombach2022high}
Robin Rombach, Andreas Blattmann, Dominik Lorenz, Patrick Esser, and Bj{\"o}rn Ommer.
\newblock High-resolution image synthesis with latent diffusion models.
\newblock In \emph{Proceedings of the IEEE/CVF conference on computer vision and pattern recognition}, pages 10684--10695, 2022.

\bibitem[Ruiz et~al.(2023)Ruiz, Li, Jampani, Pritch, Rubinstein, and Aberman]{dreambooth}
Nataniel Ruiz, Yuanzhen Li, Varun Jampani, Yael Pritch, Michael Rubinstein, and Kfir Aberman.
\newblock Dreambooth: Fine tuning text-to-image diffusion models for subject-driven generation.
\newblock In \emph{Proceedings of the IEEE/CVF conference on computer vision and pattern recognition}, pages 22500--22510, 2023.

\bibitem[Sablatnig et~al.(2002)Sablatnig, Kammerer, and Zolda]{sablatnig1998}
Robert Sablatnig, Paul Kammerer, and Ernestine Zolda.
\newblock Hierarchical classification of paintings using face- and brush stroke models.
\newblock \emph{Proc. 14th Int. Conference on Pattern Recognition}, 1, 2002.

\bibitem[Simonyan(2014)]{vgg}
Karen Simonyan.
\newblock Very deep convolutional networks for large-scale image recognition.
\newblock \emph{arXiv preprint arXiv:1409.1556}, 2014.

\bibitem[Somepalli et~al.(2024)Somepalli, Gupta, Gupta, Palta, Goldblum, Geiping, Shrivastava, and Goldstein]{somepalli2024measuring}
Gowthami Somepalli, Anubhav Gupta, Kamal Gupta, Shramay Palta, Micah Goldblum, Jonas Geiping, Abhinav Shrivastava, and Tom Goldstein.
\newblock Measuring style similarity in diffusion models.
\newblock \emph{arXiv preprint arXiv:2404.01292}, 2024.

\bibitem[Song et~al.(2020)Song, Meng, and Ermon]{ddim}
Jiaming Song, Chenlin Meng, and Stefano Ermon.
\newblock Denoising diffusion implicit models.
\newblock \emph{arXiv preprint arXiv:2010.02502}, 2020.

\bibitem[Song et~al.(2024)Song, Huang, Yao, Ye, Ci, Liu, Zhang, and Shou]{processpainter}
Yiren Song, Shijie Huang, Chen Yao, Xiaojun Ye, Hai Ci, Jiaming Liu, Yuxuan Zhang, and Mike~Zheng Shou.
\newblock Processpainter: Learn painting process from sequence data.
\newblock \emph{arXiv preprint arXiv:2406.06062}, 2024.

\bibitem[Tang et~al.(2023)Tang, Jia, Wang, Phoo, and Hariharan]{dift}
Luming Tang, Menglin Jia, Qianqian Wang, Cheng~Perng Phoo, and Bharath Hariharan.
\newblock Emergent correspondence from image diffusion.
\newblock \emph{Advances in Neural Information Processing Systems}, 36:\penalty0 1363--1389, 2023.

\bibitem[Team(2023)]{referenceonly}
Hugging~Face Team.
\newblock Stable diffusion reference implementation, 2023.
\newblock Available online.

\bibitem[Tian et~al.(2024)Tian, Wang, Zhang, and Bo]{emo}
Linrui Tian, Qi Wang, Bang Zhang, and Liefeng Bo.
\newblock Emo: Emote portrait alive-generating expressive portrait videos with audio2video diffusion model under weak conditions.
\newblock \emph{arXiv preprint arXiv:2402.17485}, 2024.

\bibitem[Wang et~al.(2024{\natexlab{a}})Wang, Spinelli, Wang, Bai, Qin, and Chen]{instantstyle}
Haofan Wang, Matteo Spinelli, Qixun Wang, Xu Bai, Zekui Qin, and Anthony Chen.
\newblock Instantstyle: Free lunch towards style-preserving in text-to-image generation.
\newblock \emph{arXiv preprint arXiv:2404.02733}, 2024{\natexlab{a}}.

\bibitem[Wang et~al.(2024{\natexlab{b}})Wang, Bai, Wang, Qin, and Chen]{instantid}
Qixun Wang, Xu Bai, Haofan Wang, Zekui Qin, and Anthony Chen.
\newblock Instantid: Zero-shot identity-preserving generation in seconds.
\newblock \emph{arXiv preprint arXiv:2401.07519}, 2024{\natexlab{b}}.

\bibitem[Wang et~al.(2023)Wang, Efros, Zhu, and Zhang]{wang2023evaluating}
Sheng-Yu Wang, Alexei~A. Efros, Jun-Yan Zhu, and Richard Zhang.
\newblock Evaluating data attribution for text-to-image models.
\newblock In \emph{ICCV}, 2023.

\bibitem[Xie et~al.(2024)Xie, Xu, Song, Wang, Shi, and Luo]{x-portrait}
You Xie, Hongyi Xu, Guoxian Song, Chao Wang, Yichun Shi, and Linjie Luo.
\newblock X-portrait: Expressive portrait animation with hierarchical motion attention.
\newblock In \emph{ACM SIGGRAPH 2024 Conference Papers}, pages 1--11, 2024.

\bibitem[Xu et~al.(2024)Xu, Zhang, Liew, Yan, Liu, Zhang, Feng, and Shou]{magicanimate}
Zhongcong Xu, Jianfeng Zhang, Jun~Hao Liew, Hanshu Yan, Jia-Wei Liu, Chenxu Zhang, Jiashi Feng, and Mike~Zheng Shou.
\newblock Magicanimate: Temporally consistent human image animation using diffusion model.
\newblock In \emph{Proceedings of the IEEE/CVF Conference on Computer Vision and Pattern Recognition}, pages 1481--1490, 2024.

\bibitem[Ye et~al.(2023{\natexlab{a}})Ye, Zhang, Liu, Han, and Yang]{ipa}
Hu Ye, Jun Zhang, Sibo Liu, Xiao Han, and Wei Yang.
\newblock Ip-adapter: Text compatible image prompt adapter for text-to-image diffusion models.
\newblock \emph{arXiv preprint arXiv:2308.06721}, 2023{\natexlab{a}}.

\bibitem[Ye et~al.(2023{\natexlab{b}})Ye, Zhang, Liu, Han, and Yang]{ye2023ip}
Hu Ye, Jun Zhang, Sibo Liu, Xiao Han, and Wei Yang.
\newblock Ip-adapter: Text compatible image prompt adapter for text-to-image diffusion models.
\newblock \emph{arXiv preprint arXiv:2308.06721}, 2023{\natexlab{b}}.

\bibitem[Zhan et~al.(2023)Zhan, Zheng, Xie, and Zisserman]{general}
Guanqi Zhan, Chuanxia Zheng, Weidi Xie, and Andrew Zisserman.
\newblock A general protocol to probe large vision models for 3d physical understanding.
\newblock In \emph{The Thirty-eighth Annual Conference on Neural Information Processing Systems}, 2023.

\bibitem[Zhang and Agrawala(2023)]{controlnet}
Lvmin Zhang and Maneesh Agrawala.
\newblock Adding conditional control to text-to-image diffusion models.
\newblock \emph{arXiv preprint arXiv:2302.05543}, 2023.

\bibitem[Zhang et~al.(2018)Zhang, Isola, Efros, Shechtman, and Wang]{lpips}
Richard Zhang, Phillip Isola, Alexei~A Efros, Eli Shechtman, and Oliver Wang.
\newblock The unreasonable effectiveness of deep features as a perceptual metric.
\newblock In \emph{Proceedings of the IEEE conference on computer vision and pattern recognition}, pages 586--595, 2018.

\bibitem[Zhang et~al.(2023)Zhang, Wang, Zhang, Zhao, Yuan, Qin, Wang, Zhao, and Zhou]{i2vgen-xl}
Shiwei Zhang, Jiayu Wang, Yingya Zhang, Kang Zhao, Hangjie Yuan, Zhiwu Qin, Xiang Wang, Deli Zhao, and Jingren Zhou.
\newblock I2vgen-xl: High-quality image-to-video synthesis via cascaded diffusion models.
\newblock \emph{arXiv preprint arXiv:2311.04145}, 2023.

\bibitem[Zhang et~al.(2024{\natexlab{a}})Zhang, Song, Liu, Wang, Yu, Tang, Li, Tang, Hu, Pan, et~al.]{ssr}
Yuxuan Zhang, Yiren Song, Jiaming Liu, Rui Wang, Jinpeng Yu, Hao Tang, Huaxia Li, Xu Tang, Yao Hu, Han Pan, et~al.
\newblock Ssr-encoder: Encoding selective subject representation for subject-driven generation.
\newblock In \emph{Proceedings of the IEEE/CVF Conference on Computer Vision and Pattern Recognition}, pages 8069--8078, 2024{\natexlab{a}}.

\bibitem[Zhang et~al.(2024{\natexlab{b}})Zhang, Song, Yu, Pan, and Jing]{fast}
Yuxuan Zhang, Yiren Song, Jinpeng Yu, Han Pan, and Zhongliang Jing.
\newblock Fast personalized text to image synthesis with attention injection.
\newblock In \emph{ICASSP 2024-2024 IEEE International Conference on Acoustics, Speech and Signal Processing (ICASSP)}, pages 6195--6199. IEEE, 2024{\natexlab{b}}.

\bibitem[Zhang et~al.(2024{\natexlab{c}})Zhang, Wei, Zhang, Song, Liu, Li, Tang, Hu, and Zhao]{stablemakeup}
Yuxuan Zhang, Lifu Wei, Qing Zhang, Yiren Song, Jiaming Liu, Huaxia Li, Xu Tang, Yao Hu, and Haibo Zhao.
\newblock Stable-makeup: When real-world makeup transfer meets diffusion model.
\newblock \emph{arXiv preprint arXiv:2403.07764}, 2024{\natexlab{c}}.

\bibitem[Zhang et~al.(2024{\natexlab{d}})Zhang, Zhang, Song, and Liu]{stablehair}
Yuxuan Zhang, Qing Zhang, Yiren Song, and Jiaming Liu.
\newblock Stable-hair: Real-world hair transfer via diffusion model.
\newblock \emph{arXiv preprint arXiv:2407.14078}, 2024{\natexlab{d}}.

\end{thebibliography}
}

\clearpage
\setcounter{page}{1}
\maketitlesupplementary

\section{Experimental Details in Different Bench}
\label{sec:rationale}

On each benchmark, the similarity scores are computed between a reference image and two candidate images, one of which is closer to the reference image. Image pair with the higher score is selected as the choice of current evaluated model. In this section, we will explain details of reference and candidate images selection for each benchmark.

\subsection{NIGHTS Dataset}
NIGHTS (Novel Image Generations with Human-Tested Similarities) is a dataset comprising 20,019 image triplets with human scores of perceptual similarity. Each triplet consists of a reference image and two distortions. This paper utilizes the test set of NIGHTS, which includes 2,120 image triplets. We calculate the DiffSim score for the reference image and the two distortions separately, using human evaluation results as the ground truth.

\subsection{Dreambench++ Dataset}

The Dreambench++ Dataset consists of generated images created using different generation methods, along with human-rated scores for how similar each image is to the original. In our experiment, we use the original image as the reference and randomly select two generated images based on it. The one with the higher human rating is considered closer to the reference. The dataset includes a total of 937 triplets.

\subsection{CUTE Dataset}

The CUTE Dataset includes photos of various instances taken under different lighting and positional conditions. In our experiment, for each category, we repeat the process 10 times: randomly selecting two images of the same instance under the same lighting and one image of a different instance under the same lighting. The two images of the same instance are considered more similar. The dataset contains a total of 1,800 triplets for comparison.

\subsection{IP Bench}

IP Bench contains 299 character classes, each with an original image and six variations generated using different consistency weights. In our experiment, we repeat the process for 5 times: using the original image as the reference and randomly selecting two generated images from the same class. The image with the higher consistency weight is considered closer to the reference. There are a total of 1,495 triplets for comparisons.

\subsection{TID2013 Dataset}

The TID2013 dataset contains 25 reference images, each distorted using 24 types of distortions at 5 different levels. In our experiment, we use a reference image as the starting point and randomly select two distorted images using the same type of distortions from the same reference. The image with a lower distortion level is considered closer to the reference. There are a total of 600 triplets for evaluation.

\subsection{Sref Dataset}

The Sref bench includes 508 styles manually selected by artists and generated by Midjounery, with each style featuring four images. When constructing image triplets, we randomly select two images from the same style and one image from a different style. We fix the random seed to construct 2,000 image triplets for quantitative evaluation.

\subsection{InstantStyle Bench}
The InstantStyle bench includes 30 styles, with each style comprising five images. When constructing image triplets, we randomly select two images from the same style and one image from a different style. We fix the random seed to construct 2,000 image triplets for quantitative evaluation.

\subsection{TikTok Dataset}

For tiktok dataset, we extract 10 frames from each video, and calculate the variance of different similarity metric scores between the first frame and other frame. A lower variance indicates that the metric demonstrates better robustness to changes in the movements of characters in the video.

\begin{table*}[ht]
\centering
\caption{Performance of diffsim across various benchmarks with different pre-trained models. Best results are highlighted in bold.}
\label{tab:5}
\small 
\begin{tabular}{@{}c|cc|cc|c|cc@{}}
\toprule
\textbf{Model / Benchmark} & \multicolumn{2}{c|}{\textbf{Human-align Similarity}} & \multicolumn{2}{c|}{\textbf{Instance Similarity}} & \multicolumn{1}{c|}{\textbf{Low-level Similarity}} & \multicolumn{2}{c}{\textbf{Style Similarity}} \\ 
 & \textbf{NIGHTS} & \textbf{Dreambench++} & \textbf{CUTE} & \textbf{IP} & \textbf{TID2013} & \textbf{Sref}  & \textbf{InstantStyle bench} \\ \midrule
DiffSim-S SD1.5                     & \textbf{86.52\%} & \textbf{71.50}\% & 72.06\% & \textbf{92.04\%} & \textbf{94.17\%} & \textbf{97.40\%} & \textbf{99.05\%} \\
DiffSim-C SD1.5                      & 79.16\% & 67.45\% & \textbf{76.17}\% & 77.06\% & 94.00\% & 94.70\% & 95.10\% \\
DiffSim-S SD-XL                      & 78.05\% & 63.93\% & 69.94\% & 83.41\% & 91.33\% & 93.05\% & 96.55\% \\
DiffSim DIT-XL/2 256                 & 63.38\% & 57.52\% & 53.44\% & 82.81\% & 83.50\% & 77.00\% & 80.15\% \\
DiffSim DIT-XL/2 512                 & 67.92\% & 57.31\% & 57.22\% & 81.00\% & 88.67\% & 78.20\% & 79.40\% \\
\bottomrule
\end{tabular}
\end{table*}

\section{Exploring Different Model Architectures}

In Table ~\ref{tab:5}, we present the performance differences of DiffSim using pre-trained models with different architectures. DiffSim-S SD1.5 leads in all benchmarks except for the CUTE dataset. DiffSim-C SD1.5 performs better on the CUTE dataset, possibly because the cross-attention layers in the U-Net architecture are particularly effective at distinguishing the subject. On the other hand, DiffSim-C uses IP-Adapter Plus, and the CLIP image encoder may become a performance bottleneck in other benchmarks. Models with higher resolution, such as SD-XL and DIT-XL/2 512, do not show performance improvement compared to lower resolution models like SD1.5 and DIT-XL/2 256. Furthermore, the performance of models using DIT as the pre-trained model is worse than using U-Net, with two possible reasons: 1. DIT splits the image into patches and then serializes them, which may lead to the loss of spatial information, which is detrimental to DiffSim, despite the use of positional encoding. 2. DIT is trained on the ImageNet dataset, which is much smaller than the SD1.5 and SD-XL models' training datasets.

\section{Additional Experimental Results}
\begin{figure}[htp]
    \centering
    \includegraphics[width=1.0\linewidth]{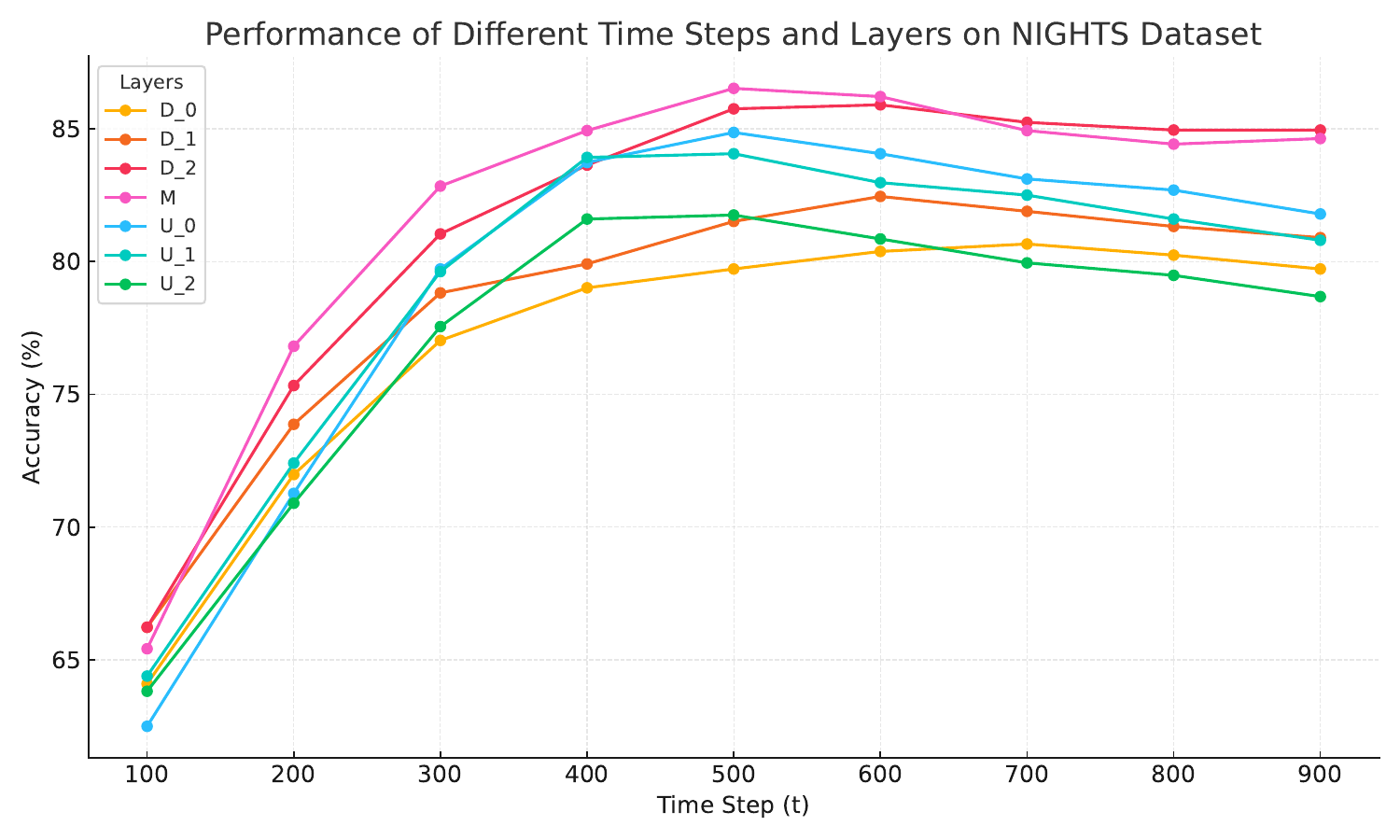}
    \caption{Results on NIGHTS dataset.}
    \label{nightsbench}
\end{figure}

\begin{figure}[htp]
    \centering
    \includegraphics[width=1.0\linewidth]{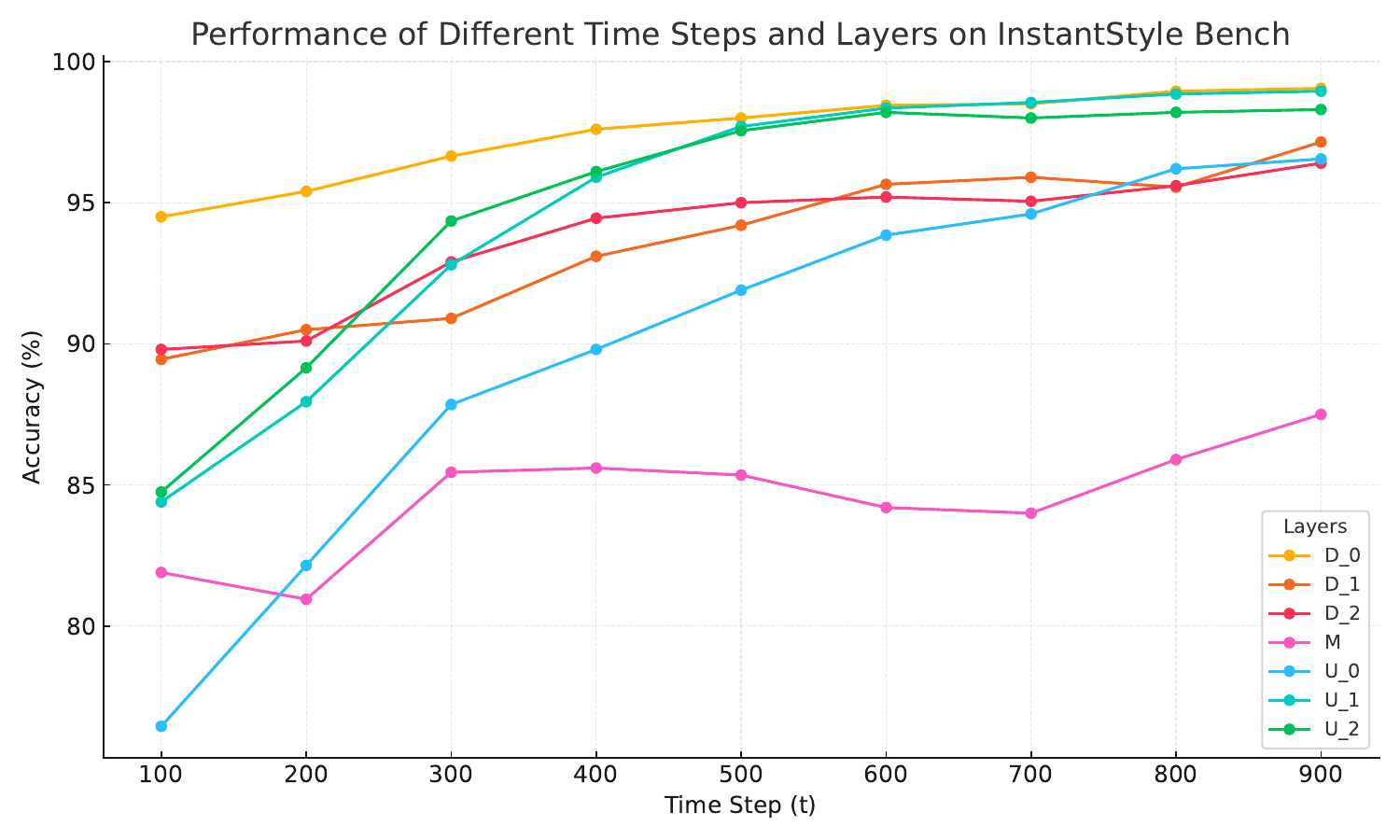}
    \caption{Results on InstantStyle bench.}
    \label{instantstylebench}
\end{figure}

In Figures \ref{nightsbench} to \ref{instantstylebench} , we present the default implementation of DiffSim, which is based on the self-attention layers of SD1.5, showing results across different layers and denoising time steps t.
% 更多实验结果
% 图7-图12，我们展示DiffSim的默认实现，即基于 SD1.5的self attention层中的实现， 不同层和去噪时间步t的结果。
% Additional Experimental Results
% In Figures 7 to 12, we present the default implementation of DiffSim, which is based on the self-attention layers of SD1.5, showing results across different layers and denoising time steps t.

\section{Additional Visual Examples}
\begin{figure*}[htp]
    \centering
    \includegraphics[width=0.89\linewidth]{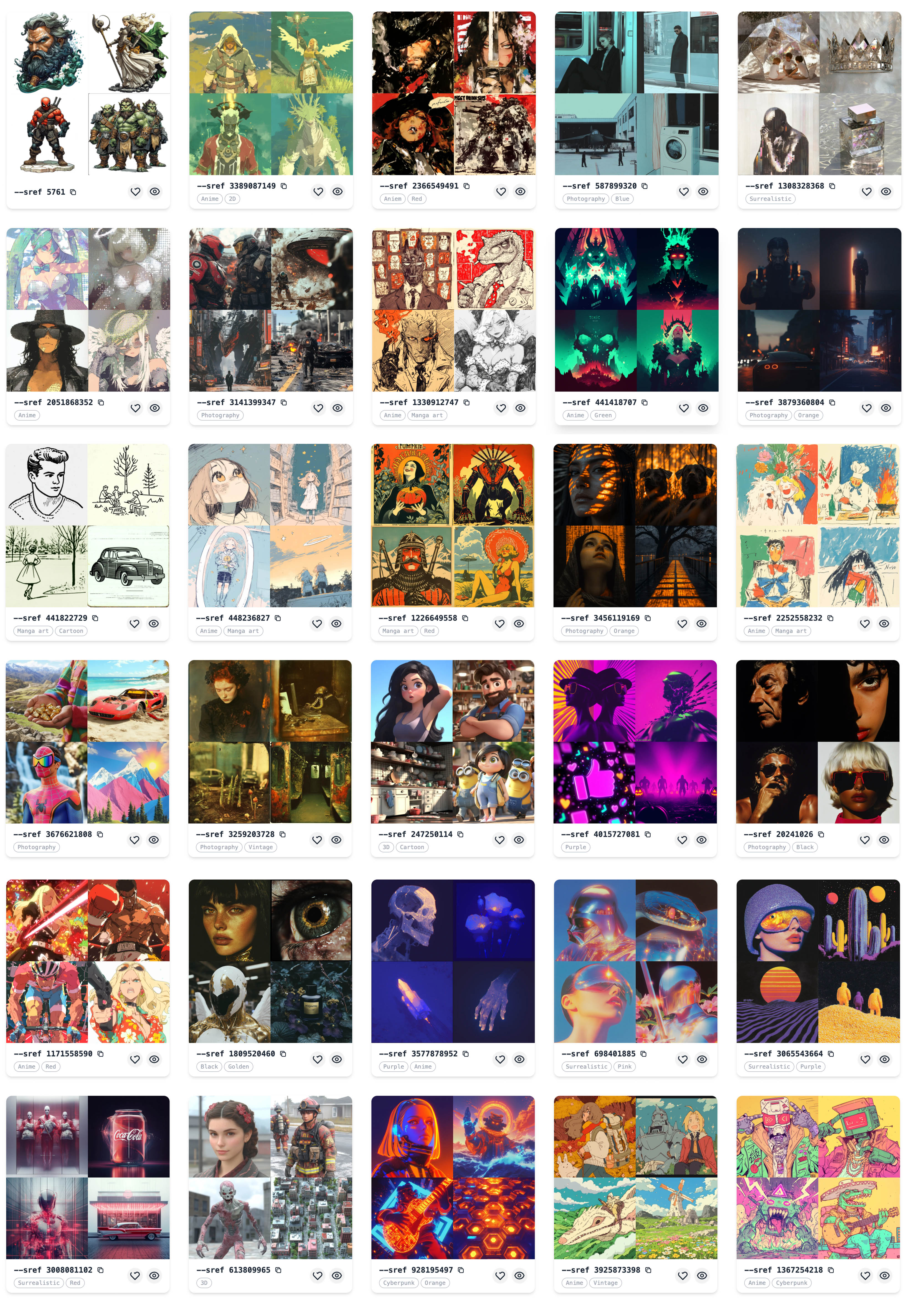}
    \caption{Examples in Sref bench we proposed.}
    \label{supp:bench}
\end{figure*}

\begin{figure*}[htp]
    \centering
    \includegraphics[width=0.89\linewidth]{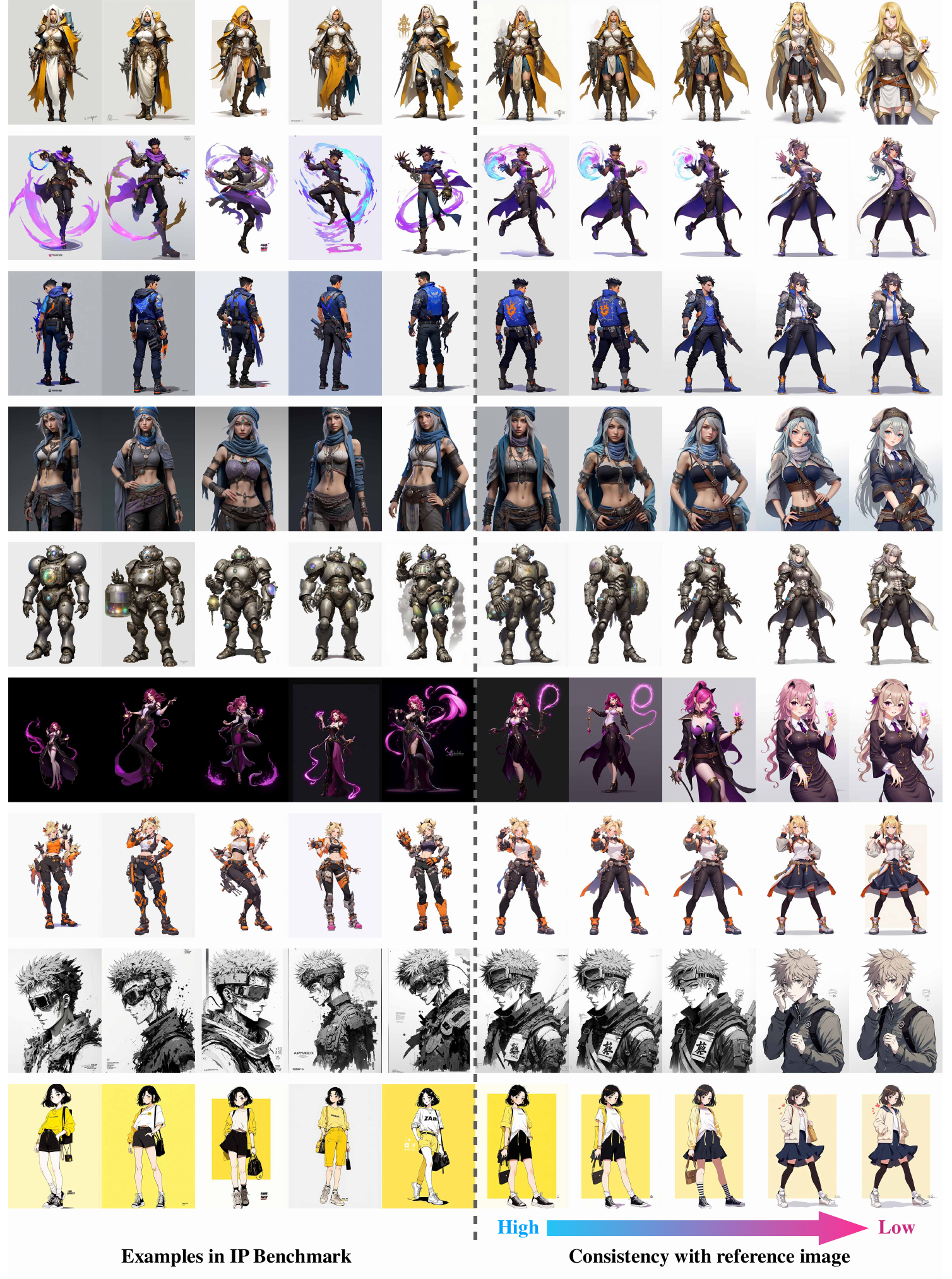}
    \caption{Examples in IP bench we proposed.}
    \label{supp:IP}
\end{figure*}

\begin{figure*}[htp]
    \centering
    \includegraphics[width=0.99\linewidth]{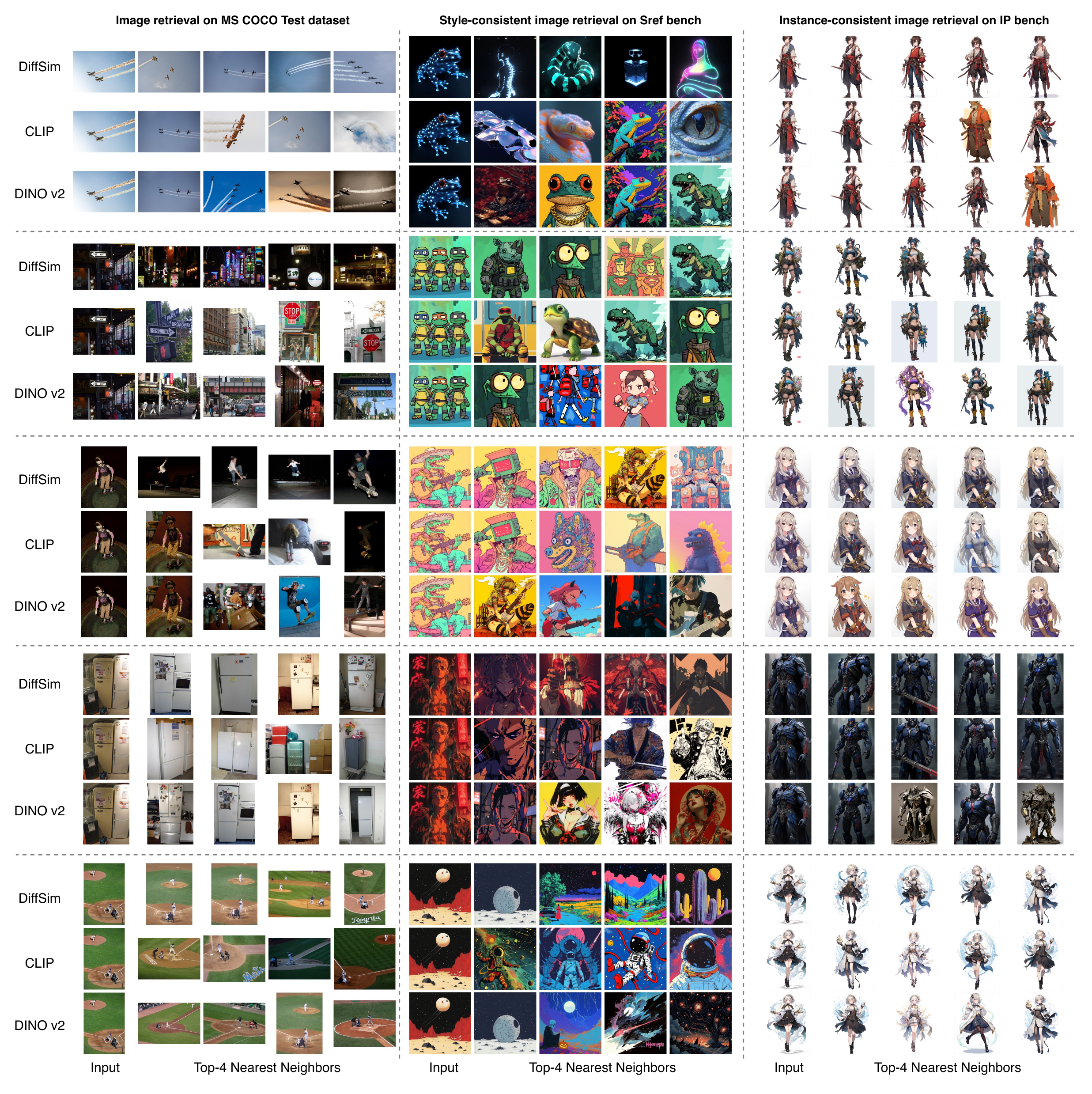}
    \caption{More image retrieval results.}
    \label{supp:retrieval}
\end{figure*}

Figure ~\ref{supp:bench} and ~\ref{supp:IP} show more examples of images from Sref bench and IP bench; Figure ~\ref{supp:retrieval} presents more top-4 retrieval results of DiffSim, CLIP, DINO v2 on MS COCO, Sref bench and IP bench.

\begin{figure}[htp]
    \centering
    \includegraphics[width=1.0\linewidth]{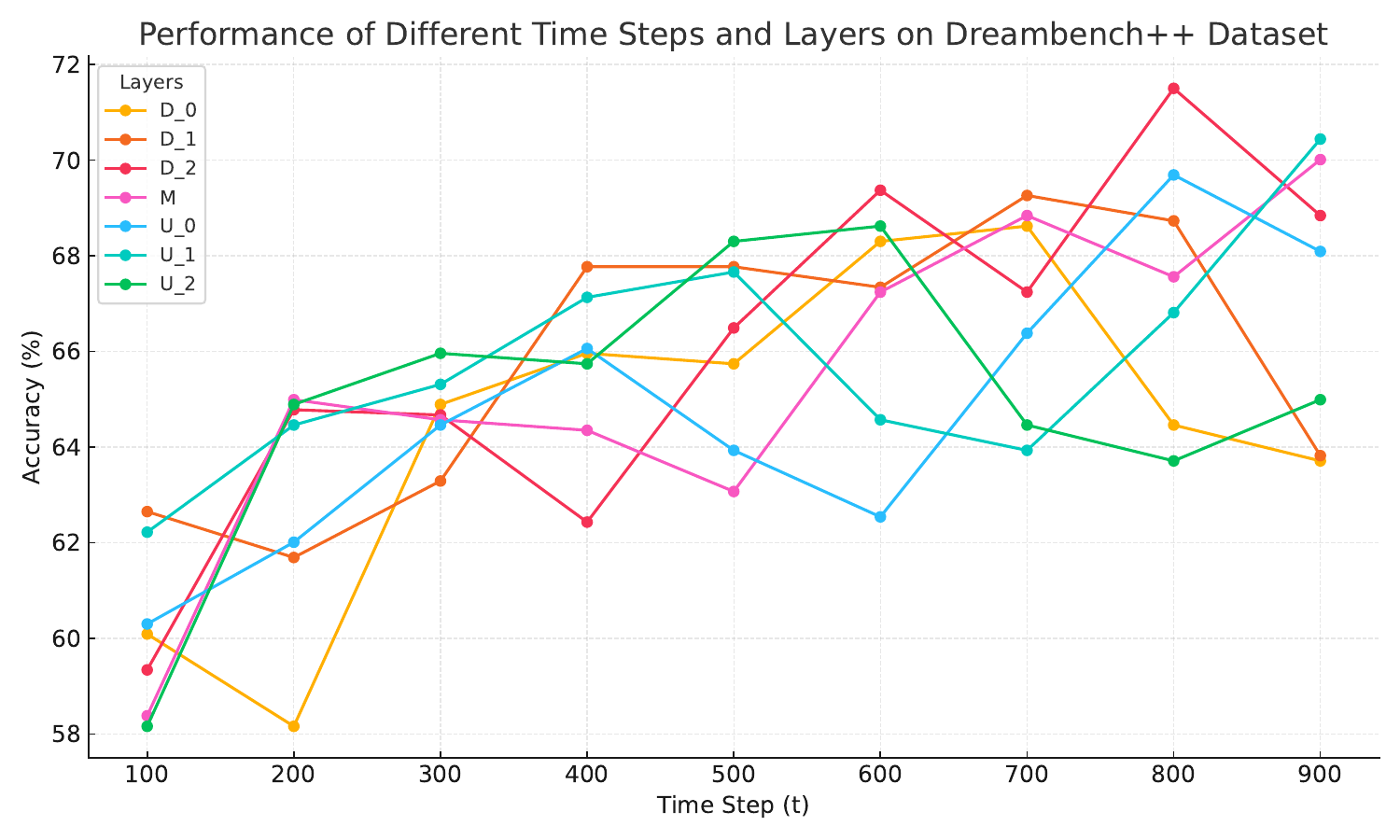}
    \caption{Results on Dreambench++ dataset.}
    \label{bench}
\end{figure}

\begin{figure}[htp]
    \centering
    \includegraphics[width=1.0\linewidth]{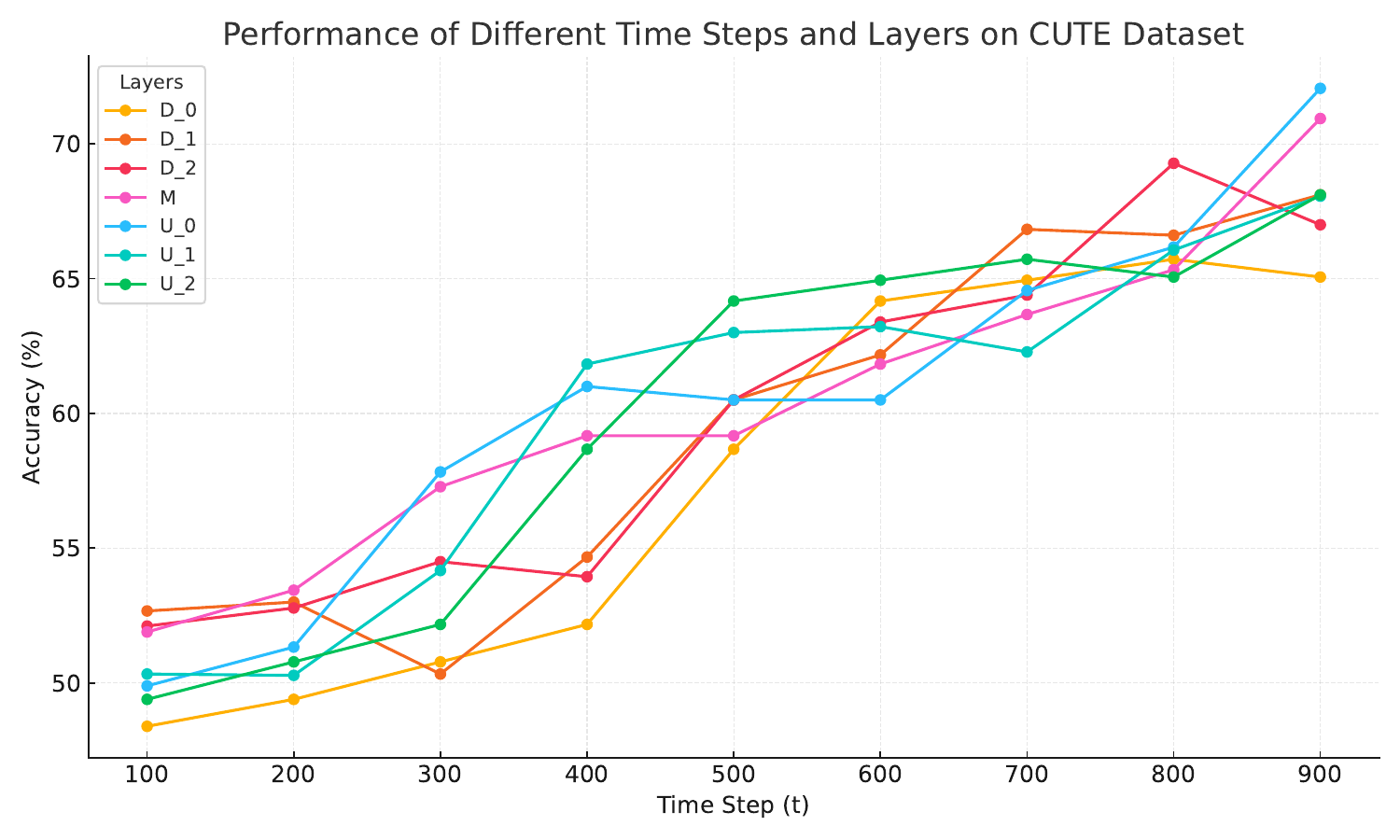}
    \caption{Results on CUTE dataset.}
    \label{bench}
\end{figure}

\begin{figure}[htp]
    \centering
    \includegraphics[width=1.0\linewidth]{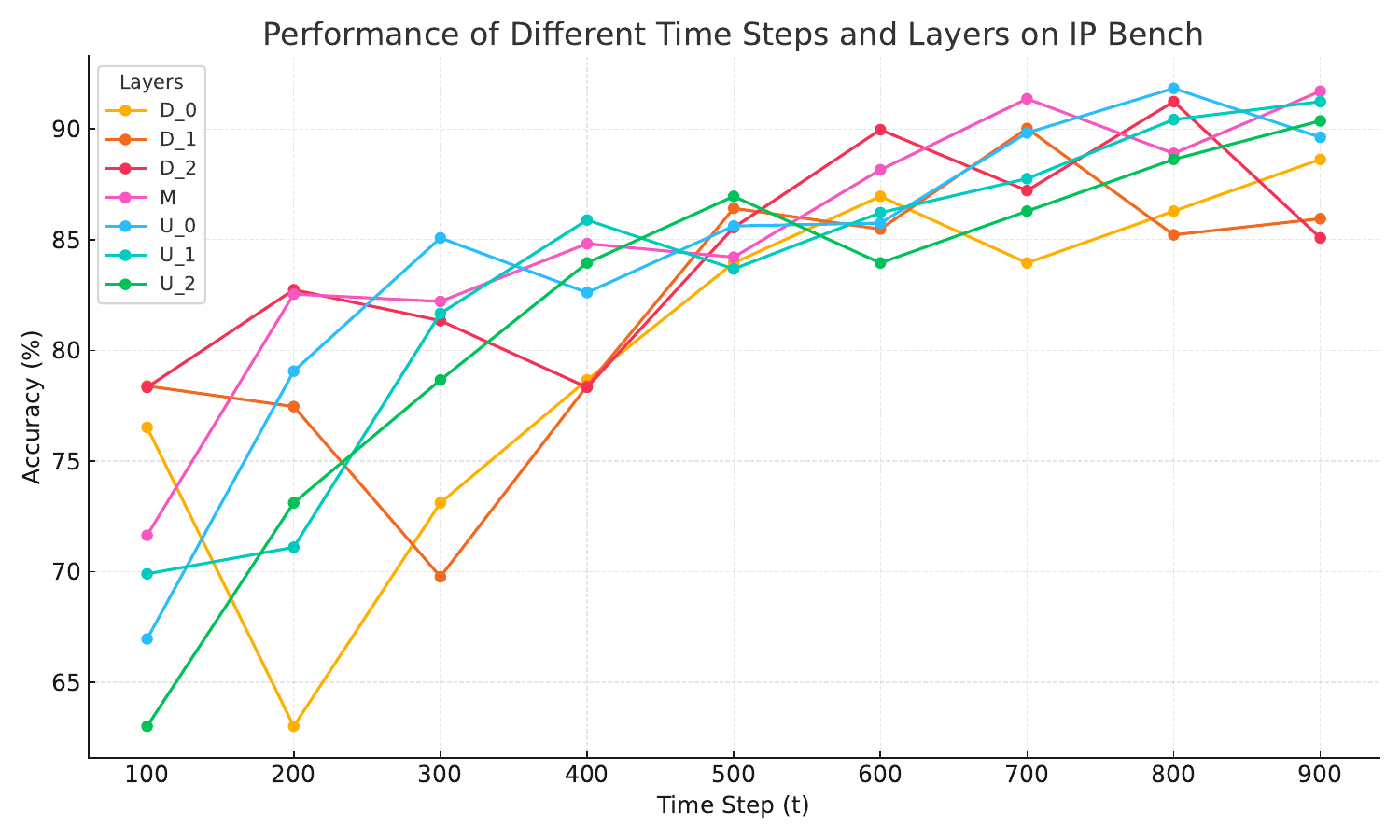}
    \caption{Results on IP bench.}
    \label{bench}
\end{figure}

\begin{figure}[htp]
    \centering
    \includegraphics[width=1.0\linewidth]{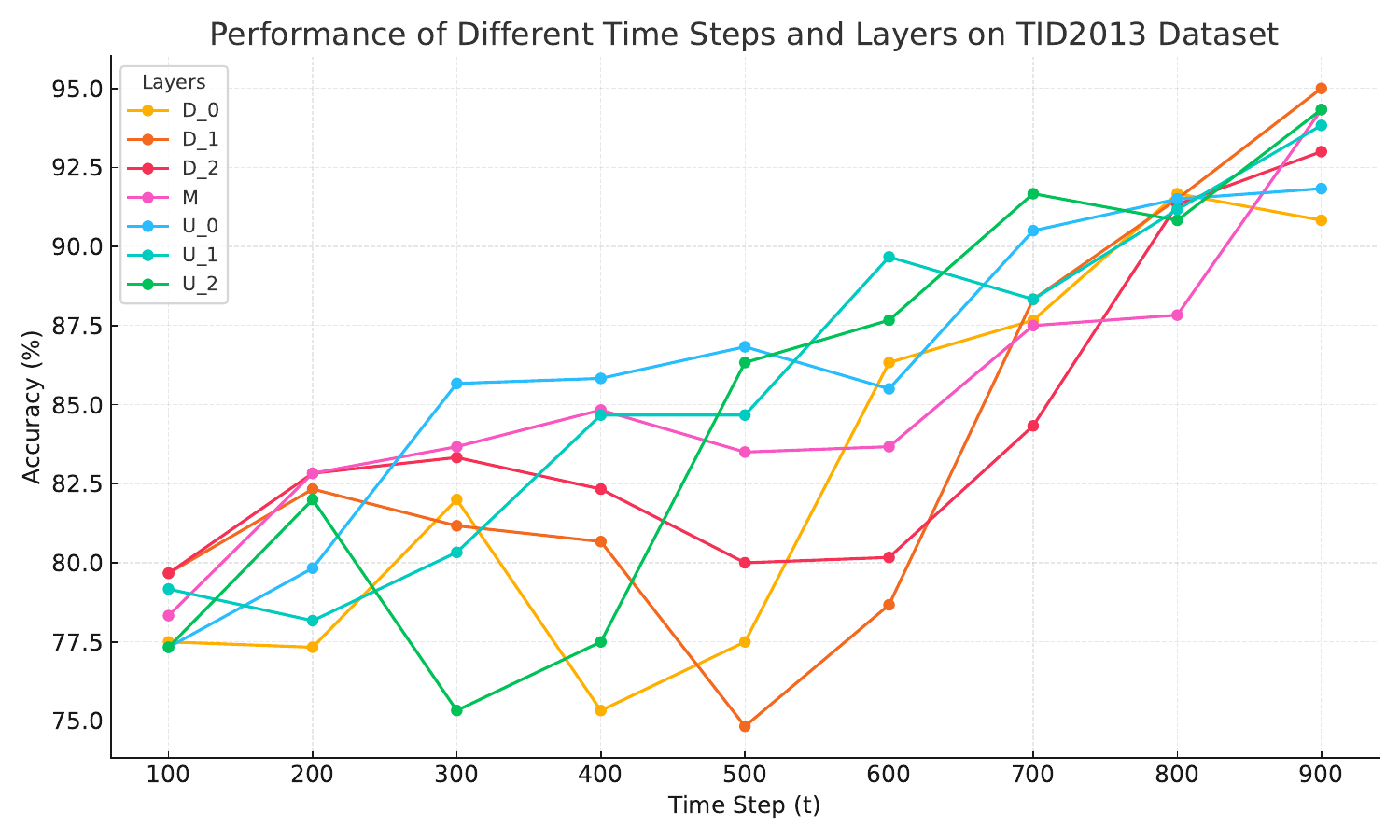}
    \caption{Results on TID2013 dataset.}
    \label{bench}
\end{figure}

\begin{figure}[htp]
    \centering
    \includegraphics[width=1.0\linewidth]{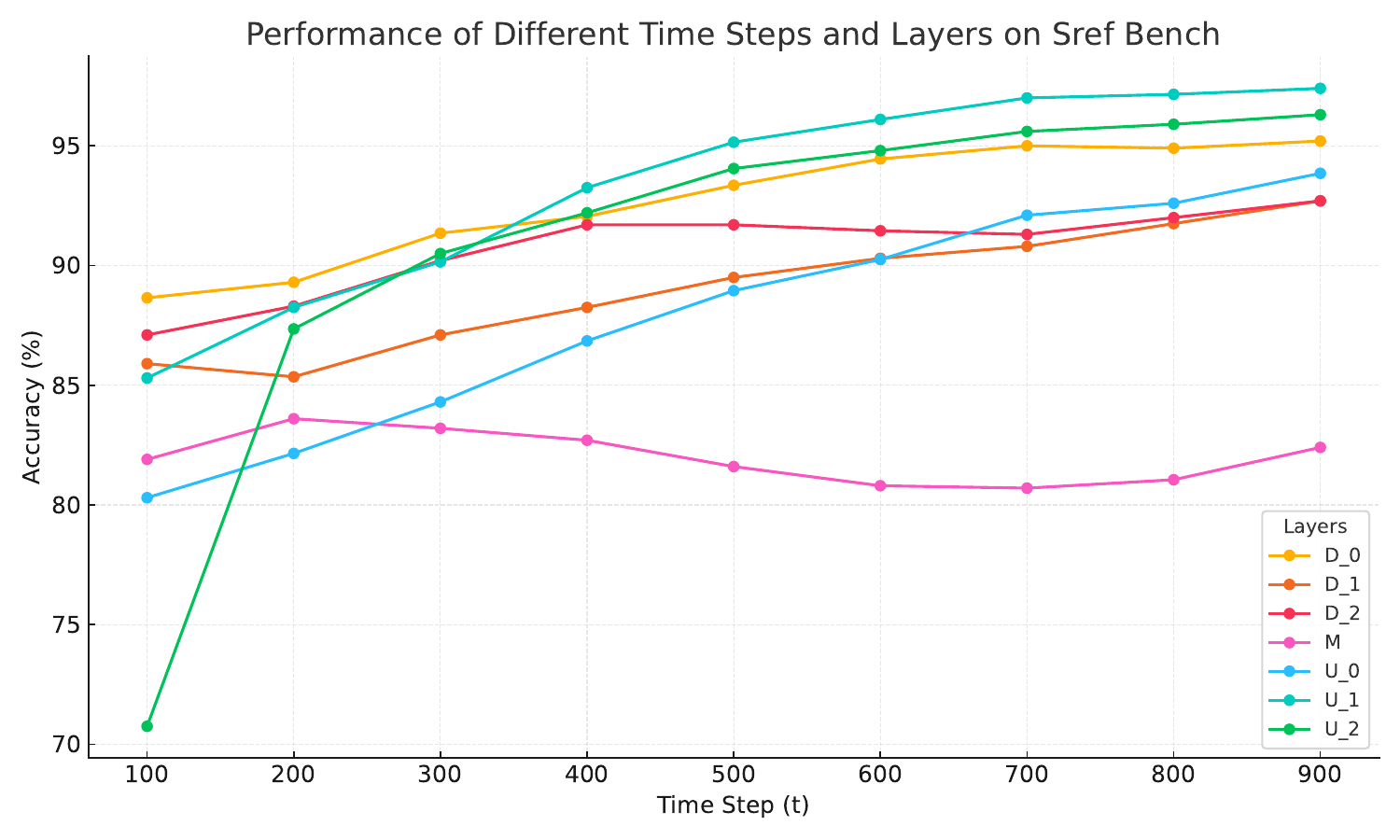}
    \caption{Results on Sref bench.}
    \label{bench}
\end{figure}

\end{document}